\documentclass[11pt,letterpaper,logo]{style}

\usepackage{bm}
\usepackage{siunitx}
\usepackage{makecell}
\usepackage{xspace}
\usepackage{placeins}
\usepackage{needspace}
\usepackage{fontawesome5}
\usepackage{longtable}
\usepackage{multicol}
\usetikzlibrary{arrows.meta}
\DeclareMathSizes{8.2}{8.2}{6}{5}
\RequirePackage{booktabs}
\RequirePackage{siunitx}
\RequirePackage{xcolor}
\RequirePackage{colortbl}
\RequirePackage{makecell}
\definecolor{MCP}{HTML}{D55E3A}
\definecolor{CLI}{HTML}{376DAA}
\definecolor{PaperText}{HTML}{22252A}
\definecolor{PaperSecondary}{HTML}{4F5964}
\definecolor{PaperGray}{HTML}{858D96}
\definecolor{Benefit}{HTML}{2E7D46}
\definecolor{Unfavourable}{HTML}{C33D36}
\definecolor{PaperLight}{HTML}{F7F8FA}

\newcommand{\NProbeStates}{338\xspace}

\newcommand{\NProbeResponses}{21,632\xspace}

\definecolor{MVGroup}{HTML}{EDF1F4}
\definecolor{MVMuted}{HTML}{657080}
\DeclareMathSizes{7.5}{8}{6}{5}
\DeclareMathSizes{7.2}{8}{6}{5}
\DeclareMathSizes{8.8}{9}{6}{5}
\DeclareMathSizes{8}{8}{6}{5}
\definecolor{SVPositive}{HTML}{477FA8}
\definecolor{SVNegative}{HTML}{BC6253}
\definecolor{SVZero}{HTML}{28313B}
\newcommand{\SVDelta}[2]{{\fontsize{7.8}{9}\selectfont\textcolor{#1}{\ensuremath{(#2)}}}}
\definecolor{SVNativeFavorable}{HTML}{297B53}
\definecolor{SVNativeUnfavorable}{HTML}{B63C3C}
\definecolor{SVNativeZero}{HTML}{303B48}
\DeclareMathSizes{8.6}{8.6}{6}{5}
\DeclareMathSizes{7.8}{7.8}{6}{5}


\DeclareMathSizes{8.5}{8}{6}{5}

\makeatletter
\expandafter\let\csname NC@find@L\endcsname\relax
\expandafter\let\csname NC@find@Y\endcsname\relax
\expandafter\let\csname NC@find@C\endcsname\relax
\makeatother
\newcolumntype{L}[1]{>{\raggedright\arraybackslash\hyphenpenalty=10000}p{#1}}
\newcolumntype{Y}{>{\raggedright\arraybackslash\hyphenpenalty=10000}X}
\newcolumntype{C}[1]{>{\centering\arraybackslash}p{#1}}

\newcommand{\PaperTitle}{Matched Capabilities, Different Behaviors: A Controlled Study of MCP and CLI Interfaces for Language Agents}
\title{\PaperTitle}
\runningtitle{\PaperTitle}
\PublicDate{2026-10-25}
\author{%
  {\Authfont
    Zhenning~Shi\textsuperscript{1}\quad
    Liang~Pang\textsuperscript{2}\quad
    Yongqi~Li\textsuperscript{1}\advisor
  }\\
  {\Affilfont
    \textsuperscript{1}The Hong Kong Polytechnic University\\
    \textsuperscript{2}Institute of Computing Technology, Chinese Academy of Sciences\\
    \texttt{zhenning.shi@connect.polyu.hk, liyongqi0@gmail.com}
  }
}
\hypersetup{pdftitle={\PaperTitle},pdfauthor={Zhenning Shi, Liang Pang, Yongqi Li},pdfsubject={}}
\newcommand{\CodeAvailabilityStatement}{The code is available in our \href{https://github.com/ModalityDance/MICA}{repository}.}
\newcommand{\TitleLinks}{%
\centering
    \vspace{6pt}
    {\noindent\absfont\fontsize{11}{13}\selectfont\urlstyle{tt}%
    \faGithub\ GitHub: \url{https://github.com/ModalityDance/MICA}\par}%
}

\begin{document}
\raggedbottom
\begin{abstract}

Language agents increasingly act through external tools and interactive
environments, making the interface between the model and its executable
capabilities part of the decision process. Model Context Protocol (MCP) and
command-line interfaces (CLI) can expose the same semantic operations through
markedly different model-facing representations. We study whether these
representational differences change agent behavior when executable capabilities
are matched.
We introduce Matched Interface Comparison and Analysis (MICA), which
renders MCP and CLI from a shared semantic action contract and routes decoded
actions through the same runtime. Across six models and three interactive
environments, matched interfaces produce substantial differences in task
performance and interaction cost. These effects extend beyond surface validity
to semantic action selection, failure pathways, and recovery. Matched-state
probes localize sensitivity to capability presentation, realization, and
feedback. Paired natural trajectories reveal structured divergence in 94.3\%
of 2,004 episodes and show that early differences propagate through subsequent
interaction.
This analysis identifies interface realization as a direct intervention
boundary. We therefore introduce Decoupled Action Realization and
Targeted Repair (DART), which preserves validated semantic commitments while
repairing interface-specific realization. DART improves benchmark score in all
six evaluated settings by up to 9.0 points and reduces MCP token cost in all
three environments by up to 27.1k tokens per task. Together, these results
establish interface representation as part of the closed-loop decision
environment shaping agent behavior.

\end{abstract}
\maketitle

\section{Introduction}

Modern language agents increasingly act through tools, APIs, and executable
environments
\citep{yao2023react,toolsandbox,li2023apibank,liu2024agentbench}.
These external capabilities are accessed through interfaces that specify how
available actions, arguments, response conventions, and execution feedback are
presented to the model. In an interactive agent, however, an interface does more
than expose what actions are possible: it becomes part of the context from which
the next action is selected. As a result, different representations of the same
executable capabilities can induce different local decisions and, through
closed-loop interaction, different trajectories. Prior work has shown that
action representations~\citep{wang2024codeact}, tool metadata
\citep{yuan2025easytool,blankenstein2026biasbusters}, and agent--computer
interfaces~\citep{yang2024sweagent} materially shape tool use and task
performance. This raises a fundamental question for interactive language
agents: \emph{When executable capabilities are held fixed, to what extent does
their model-facing representation change the policy that emerges?}

This question is particularly important because both structured tool invocation
and terminal-based interaction are increasingly common in modern agent systems.
The Model Context Protocol (MCP) presents capabilities as named tools with
schema-defined arguments
\citep{anthropic2024mcp,mcp2026tools}, whereas a command-line interface (CLI)
expresses operations through commands, flags, and textual conventions. These
interfaces can encode the same semantic operations while presenting
systematically different cues for selecting and expressing them. Both MCP and
CLI are now widely used in modern agent systems
\citep{anthropic2025claudecode,anthropic2025claudecodemcp,
google2025geminicli,github2025copilotcli,openai2025codexcli}, making a
systematic study of how their representational differences affect agent
behavior increasingly important. MCP and CLI therefore provide a natural
setting for disentangling \emph{what an agent can do} from \emph{how those
capabilities are represented to the model}.

To answer this question, we need a controlled comparison that holds executable
semantics fixed while varying only the model-facing representation. To make
such a comparison informative, however, we must solve two attribution problems.
First, system-level MCP--CLI comparisons often change more than the
interface. Tool granularity~\citep{bechard2026terminal}, prompting, public
information, validation, or execution scaffolding can change at the same time
\citep{alier2026scaffolding,xu2026devil}. A performance difference under such
conditions cannot be attributed to representation alone. Second, even after
executable semantics are matched, endpoint metrics collapse distinct
interaction pathways~\citep{wang2026mcpbench}. The same final score can arise
from different semantic choices, realization failures, and
interaction costs, while an early local difference can alter the history and
state conditioning all later decisions. Systematically comparing how MCP and
CLI representations shape agent behavior therefore requires both
\emph{semantic matching} and \emph{trajectory-level attribution}.

We introduce Matched Interface Comparison and Analysis (MICA) to systematically
compare MCP and CLI under matched executable semantics. By holding the
underlying executable capabilities fixed, MICA attributes behavioral differences
to model-facing representation rather than differences in what the agent can do.
It then analyzes interface effects at two complementary scales. At fixed decision
contexts, \emph{matched-state probes} vary capability presentation, realization,
and feedback to localize where interface sensitivity enters the policy. Along
\emph{paired natural trajectories}, MICA identifies the first consequential
divergence and traces how that local difference propagates through subsequent
history, state, recovery, cost, and task outcome.
This analysis further separates semantic action selection from
interface-specific realization and identifies realization as a direct
intervention boundary. A semantic action may already be valid even when its
interface-specific expression fails. Guided by this analysis, we introduce
Decoupled Action Realization and Targeted Repair (DART) to address
interface-specific realization failures without discarding an already
validated semantic decision. Once a semantic action is validated, DART
preserves it as a fixed commitment. If its realization fails while the
commitment remains valid, DART repairs only the interface-specific expression
rather than selecting a new semantic action.

Our experiments reveal a consistent mechanistic picture. Across six models and
three interactive environments, MCP and CLI with matched executable capabilities still produce
substantial differences in task performance and interaction cost.
These differences extend beyond surface validity. Interface
representation changes semantic action selection,
while matched-state analysis shows that presentation, realization,
and feedback interact strongly in shaping local behavior. Paired trajectories
further show that consequential divergence is widespread and often emerges
early, after which interaction histories and costs separate.
Acting on the realization boundary identified by MICA, DART provides a
direct intervention on the identified realization mechanism, improving task
performance across all six evaluated benchmark--interface settings. Under MCP,
it also reduces interaction cost. Together, these
results establish interface representation as an active part of the
closed-loop decision environment through which agent policies are realized.

\Needspace{5\baselineskip}
In summary, our contributions are as follows:
\begin{itemize}
    \item We provide a study of MCP and CLI under matched executable
    capabilities, showing that interface representation can substantially alter
    agent behavior and interaction cost.

    \item We introduce Matched Interface Comparison and Analysis (MICA), which
    localizes interface sensitivity and traces how local differences propagate
    through closed-loop interaction.

    \item We introduce Decoupled Action Realization and Targeted Repair (DART),
    which repairs interface-specific realization while preserving validated semantic
    decisions, enabling more reliable recovery from the realization failures
    identified by MICA.
\end{itemize}
\section{Related Work}

\paragraph{Tool Use and Interface Representation.}
Language agents have progressed from interleaved reasoning and acting
\citep{yao2023react} toward stateful, multi-step interaction with external
tools and environments
\citep{patil2025bfcl,toolsandbox,wang2024mint,yao2025taubench}.
A growing body of work shows that agent behavior depends on how available
actions are represented to the model. Executable code, structured tool calls,
and textual actions induce different interaction patterns
\citep{wang2024codeact}. Agent--computer interface design also affects
downstream behavior \citep{yang2024sweagent}, while changes in tool names,
descriptions, ordering, and other interface metadata alter action selection
\citep{blankenstein2026biasbusters,yuan2025easytool,xu2026devil}.
Related work on constrained generation and conversational recovery further
shows that valid action realization and post-failure correction are integral
parts of tool-mediated interaction
\citep{banerjee2025crane,kim2026rein}.
Together, these studies establish interface representation as an important
source of behavioral variation and motivate controlled attribution under
matched executable semantics.

\paragraph{MCP-based Tool Use.}
The Model Context Protocol standardizes tool discovery and invocation through
named tools with schema-defined arguments
\citep{anthropic2024mcp,mcp2026tools}.
Recent benchmarks including MCP-Bench \citep{wang2026mcpbench}, MCPMark
\citep{wu2026mcpmark}, MCP-AgentBench \citep{guo2026mcpagentbench}, and
MCP-Universe \citep{luo2025mcpuniverse} evaluate increasingly complex MCP
interaction involving multi-step planning, tool discovery, and cross-tool
coordination. These benchmarks characterize agent capability within MCP-native
tool representations and provide broad coverage of realistic MCP use.
System-level comparisons of MCP and CLI across multiple agent scaffoldings
also report substantial variation across deployed configurations
\citep{alier2026scaffolding}. MICA studies a complementary question by holding
the semantic action contract and post-decoding runtime fixed while varying the
model-facing presentation, realization, and feedback components of the
interface.

\paragraph{CLI and Terminal Agents.}
Terminal environments expose actions through textual commands and iterative
execution feedback, as studied in InterCode \citep{yang2023intercode} and
Terminal-Bench \citep{merrill2026terminalbench}. These environments capture
central properties of command-line interaction including broad command
grammars, compositional actions, and long-horizon workflows. Recent
enterprise-automation work directly compares terminal agents with MCP-based
tool registries and shows that capability packaging explains a substantial
part of the observed performance difference
\citep{bechard2026terminal}. This finding makes capability matching a central
control for studying interface representation. Our construction therefore
aligns benchmark-specific semantic operations and downstream execution
semantics before comparing MCP-style tool calls with CLI-style commands.
\section{Problem Formulation}
\label{sec:setting}

\paragraph{Semantic actions and capability matching.}
We consider a language agent $\pi_\theta$ interacting with a stateful
environment. At decision step $t$, the environment is in state $s_t$, while
the agent observes a policy-visible observation $o_t$ and public interaction
history $h_t$. The benchmark defines a semantic action contract
$\mathcal{C}_t$, under which the agent selects
\[
a_t=(u_t,v_t)\in\mathcal{A}_{\mathcal{C}_t}(s_t),
\]
where $u_t$ denotes an operation and $v_t$ its normalized typed arguments.
We write $a\equiv_{\mathrm{sem}}a'$ when two proposals specify the same
registered operation with equivalent normalized arguments under the active
contract. This normalized semantic action is the common unit of comparison
across interfaces. MCP and CLI are constructed from the same
benchmark-specific contract and therefore expose matched executable
capabilities through different model-facing representations.

\paragraph{Interface factorization and shared execution.}
For $I\in\{\mathrm{MCP},\mathrm{CLI}\}$, we represent the interface as
$I=(\Phi_I,R_I,\Psi_I)$ with $R_I=(\rho_I,D_I)$. Here, $\Phi_I$ specifies
capability presentation, $\rho_I$ the response protocol, $D_I$ the
interface-specific decoder, and $\Psi_I$ the feedback representation. Given
\[
x_t^I=\Phi_I(\mathcal{C}_t,o_t,h_t),
\]
the model samples
$y_t^I\sim\pi_\theta(\cdot\mid x_t^I)$
and produces a decoded proposal
\[
\hat a_t^I
=
D_I(y_t^I)
\in
\mathcal{A}_{\mathcal{C}_t}(s_t)\cup\{\bot\},
\]
where $\bot$ denotes a response that does not decode to a registered semantic
action. Decoded proposals then enter the shared semantic validation,
native-action compilation, and execution path
\begin{equation}
    \hat a_t^I
    \xrightarrow{\;V,K\;}
    n_t
    \xrightarrow{\;E\;}
    (s_{t+1},\omega_t),
    \qquad
    z_t^I=\Psi_I(\omega_t).
    \label{eq:shared_execution}
\end{equation}
Here, $V$, $K$, and $E$ denote semantic validation, native-action compilation,
and environment execution, respectively, and $n_t$ is the compiled native
action. Model-facing variation enters through
$(\Phi_I,\rho_I,D_I,\Psi_I)$, while both interfaces share these downstream
runtime operators. This factorization distinguishes response decoding,
semantic admission, and environment execution.

\paragraph{Interaction equivalence and consequential divergence.}
Interface effects are analyzed at two complementary scales. At a matched
decision context, fixing the environment state, public history, semantic
contract, model, and generation configuration supports local comparison under
different assignments of $\Phi$, $R$, and $\Psi$. Along closed-loop
interaction, an early difference changes the observations, histories, and
environment states that condition subsequent decisions.

Let
\[
b_t^I
=
(\hat a_t^I,f_t^I,\omega_t^I,r_t^I)
\]
denote the interaction signature at step $t$, consisting of the semantic
proposal, failure stage, execution outcome, and recovery status. We write
$b_t^I\equiv b_t^J$ when the proposals are semantically equivalent and the
remaining components agree. The first consequential divergence is
\begin{equation}
    t^\star
    =
    \min\left\{
        t
        \mid
        b_t^{\mathrm{MCP}}
        \not\equiv
        b_t^{\mathrm{CLI}}
    \right\}.
    \label{eq:first_divergence}
\end{equation}
Pairs with no such step are recorded as having no observed divergence. Surface
variation remains within the common prefix whenever both responses decode to
the same semantic action and induce the same interaction outcome. Before
$t^\star$, the aligned contexts support matched interface comparison. After
$t^\star$, each episode evolves under its own history and environment state,
and subsequent differences characterize downstream propagation. This
separation provides the basis for the matched-state and trajectory analyses in
Section~\ref{sec:controlled_origins}.
\section{Matched Interface Comparison and Analysis}
\label{sec:framework}

\begin{figure}[t]
    \centering
    \includegraphics[width=\linewidth]
    {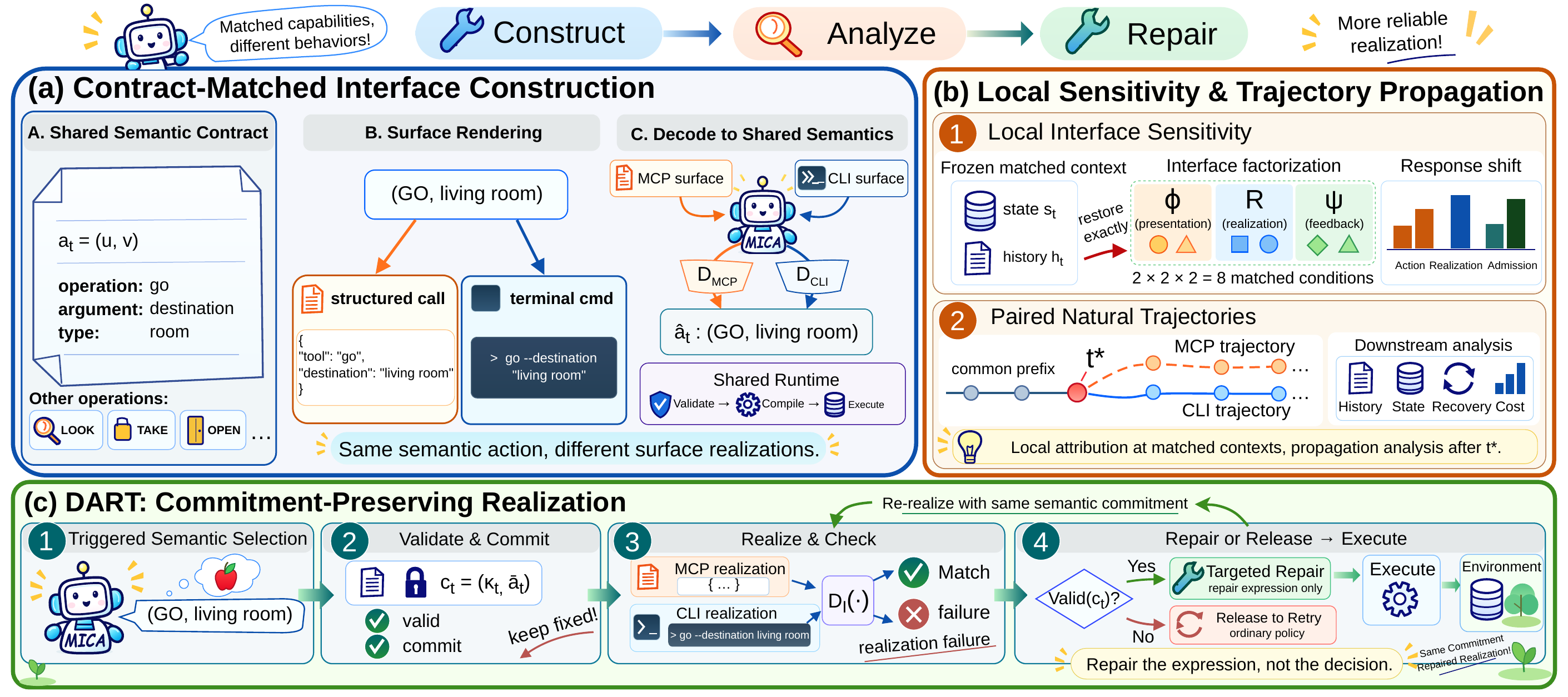}
    \caption{
    \textbf{The overview of MICA.}
    MCP and CLI render a shared semantic action contract through different
    model-facing interfaces while sharing downstream validation and execution.
    Matched-state probes localize interface sensitivity, paired trajectories
    trace its propagation, and DART preserves semantic commitments during
    realization repair.
    }
    \label{fig:framework_overview}
\end{figure}

Matched Interface Comparison and Analysis (MICA) isolates the behavioral effect
of interface representation while keeping executable capabilities fixed. It
combines contract-matched MCP--CLI construction, matched-state and trajectory
analysis, and commitment-preserving realization with DART
(Figure~\ref{fig:framework_overview}).

\subsection{Contract-Matched Interface Construction}
\label{sec:shared_actions}

\paragraph{Shared semantic contract.}
MICA constructs MCP and CLI from a common benchmark-specific semantic contract,
fixing the executable action space while varying its model-facing
representation. At decision step $t$, the contract is
\begin{equation}
    \mathcal{C}_t
    =
    \left\{
        \bigl(u,\Sigma_u,\mathcal{V}_u(s_t),\Gamma_u\bigr)
        \mid
        u\in\mathcal{U}_t
    \right\},
\label{eq:semantic_contract}
\end{equation}
where $\Sigma_u$ specifies the typed arguments of operation $u$,
$\mathcal{V}_u(s_t)$ its state-local argument domain, and $\Gamma_u$ its native
execution binding.

\paragraph{Interface rendering and shared execution.}
For $I\in\{\mathrm{MCP},\mathrm{CLI}\}$, MICA renders the contract as
$\mathcal{S}_t^I=\operatorname{Render}_I(\mathcal{C}_t)$. MCP exposes named
tools with structured arguments, whereas CLI exposes registered commands with
textual argument bindings. Both interfaces decode responses into the same
normalized semantic action space and then enter the shared validation,
compilation, and execution path in Equation~\ref{eq:shared_execution}.
Consequently, surface syntax can vary while executable semantics remain fixed.
Appendices~\ref{app:interface_conformance} and~\ref{app:action_mappings} give
the rendering rules, benchmark-specific mappings, and conformance checks.

\subsection{Local Interface Sensitivity and Trajectory Propagation}
\label{sec:controlled_origins}

MICA separates local interface sensitivity from its downstream propagation.
Matched-state probes vary interface components at a fixed decision context,
while paired natural trajectories trace what follows after the first
consequential divergence.

\paragraph{Matched-state probes.}
At a registered decision context, let
$\xi=(\xi_{\Phi},\xi_R,\xi_{\Psi})
\in\{\mathrm{MCP},\mathrm{CLI}\}^{3}$
denote the presentation, realization, and feedback forms assigned to the
model. Holding the environment state, public history, semantic contract, model,
and generation configuration fixed, each assignment induces a local response
distribution
$P_{\xi}^{\,t}(b)
=P(b_t=b\mid s_t,h_t,\mathcal{C}_t;\xi)$.
The two homogeneous assignments recover the primary MCP and CLI interfaces.
The remaining six exchange individual components under the same semantic
contract and downstream runtime.

For $k\in\{\Phi,R,\Psi\}$, let $\xi^{k\leftarrow I}$ replace component $k$
with its form under interface $I$. We summarize the component main effect for a
registered contrast $d$ as
\begin{equation}
    \Delta_k(d)
    =
    \mathbb{E}_{\xi_{-k}}
    \left[
        d\!\left(
            P_{\xi^{k\leftarrow\mathrm{MCP}}}^{\,t},
            P_{\xi^{k\leftarrow\mathrm{CLI}}}^{\,t}
        \right)
    \right].
\label{eq:component_contrast}
\end{equation}
This averages the MCP--CLI contrast for component $k$ over assignments of the
remaining components. The same $2\times2\times2$ factorial coding yields the
pairwise and three-way interaction contrasts used in the local analysis.
Sampling and estimation details are given in
Appendix~\ref{app:matched_probes}.

\paragraph{Natural trajectory propagation.}
Paired MCP and CLI episodes begin from aligned initial conditions and are
compared through the first consequential divergence $t^\star$ defined in
Equation~\ref{eq:first_divergence}. Before $t^\star$, the aligned interaction
supports matched comparison. After $t^\star$, each episode follows its own
history and environment state. We characterize this natural suffix through
semantic-action agreement, failure and recovery, interaction cost, and terminal
outcome, thereby separating local attribution from downstream propagation.
Alignment rules and suffix-level estimators are detailed in
Appendix~\ref{app:trajectory_analysis}.

\subsection{DART: Commitment-Preserving Realization}
\label{sec:dart}

DART (\textbf{D}ecoupled \textbf{A}ction \textbf{R}ealization and
\textbf{T}argeted Repair) targets the realization boundary exposed by MICA.
It separates semantic action selection from interface-specific expression so
that a validated decision remains fixed while its realization is repaired.

\paragraph{Semantic commitment and realization.}
DART first elicits a normalized semantic action
$\bar a_t=(u_t,v_t)$ under the active public contract, binds it to the contract
identity $\kappa_t$, and then realizes the resulting commitment through the
assigned interface,
\begin{equation}
    \bar a_t
    \sim
    \pi_{\theta,I}^{\mathrm{sel}}
    \!\left(\cdot\mid x_t^I\right),
    \qquad
    c_t=(\kappa_t,\bar a_t),
    \qquad
    y_t^I
    \sim
    \pi_{\theta,I}^{\mathrm{real}}
    \!\left(\cdot\mid x_t^I,c_t\right).
\label{eq:dart_pipeline}
\end{equation}
Both phases use the same underlying model with phase-specific response
contracts. Selection fixes the semantic operation and typed arguments. The
realization phase determines only their MCP or CLI expression. A realization
is admitted when it decodes to the committed action and the commitment remains
valid under the active contract. The complete preservation gate is specified
in Appendix~\ref{app:dart_implementation}.

\paragraph{Commitment-preserving repair.}
If the initial realization fails while $c_t$ remains valid, DART keeps the
commitment fixed and issues one preserving correction conditioned on the failed
expression and its public feedback. The correction may change the
interface-specific expression while preserving the committed operation and
every normalized typed argument. An admitted correction proceeds through the
shared runtime. Otherwise, control returns to the ordinary retry policy. This
intervention isolates realization repair from semantic reselection, providing a
direct test of the realization mechanism identified by MICA. Trigger scopes,
controller logic, and phase-specific prompts are detailed in
Appendices~\ref{app:dart_implementation} and~\ref{app:prompts}.
\section{Experimental Setup}
\label{sec:experiments}
\label{sec:experimental_setup}
\label{sec:experimental_matrix}
\label{sec:evaluation_protocol}

\paragraph{Benchmarks and models.}
We evaluate agents in three interactive environments with complementary
demands, namely ALFWorld~\citep{shridhar2021alfworld} for grounded household
interaction, WebShop~\citep{yao2022webshop} for goal-directed web shopping,
and ScienceWorld~\citep{wang2022scienceworld} for multi-step scientific
reasoning. Our model suite includes Qwen3-4B/8B/14B/32B
\citep{yang2025qwen3}, DeepSeek-v4-Flash~\citep{deepseekai2026deepseekv4},
and Gemini-2.5-Flash~\citep{comanici2025gemini25}. Native benchmark
interfaces are retained as calibration references.

\paragraph{Controlled evaluation.}
For each task, MCP and CLI are rendered from the same semantic action contract
(Section~\ref{sec:shared_actions}) while holding fixed the model, initial task
state, public information, semantic action space, validation and execution
pipeline, generation settings, and interaction budget. Primary metrics are
benchmark score or success and total charged model tokens per task. We further
measure semantic action selection, failure and recovery behavior, matched-state
component sensitivity, first consequential divergence, and post-divergence
trajectory behavior. For RQ4, Gemini-2.5-Flash is evaluated in all six
benchmark--interface settings against a protocol-matched Base. DART changes
only the commitment-preserving controller and registered trigger configuration.
Full task sets, budgets, estimators, and DART protocol details are given in
Appendices~\ref{app:evaluation_protocol} and~\ref{app:dart_implementation}.

\paragraph{Research questions.}
We organize the evaluation around four research questions.
\textbf{RQ1.} How does interface representation affect task performance and
interaction cost under matched executable capabilities?
\textbf{RQ2.} How does interface representation alter semantic decisions and
recovery behavior?
\textbf{RQ3.} Where do interface-induced differences originate, and how do
they propagate through subsequent interaction?
\textbf{RQ4.} How does commitment-preserving realization affect task
performance and interaction cost after realization failures?
\section{Results}
\label{sec:results}

We report the results in the order of the four research questions, moving from
task-level performance and interaction cost to failure and recovery, local
interface effects and trajectory propagation, and finally the DART
intervention.

\subsection{Task Performance and Interaction Cost}
\label{sec:rq1}

\begin{table}[!htb]
    \centering
    \begingroup\centering\fontsize{8}{9.6}\selectfont
\setlength{\tabcolsep}{2pt}\renewcommand{\arraystretch}{1.13}
\setlength{\aboverulesep}{2pt}\setlength{\belowrulesep}{2pt}
\begin{tabular*}{\linewidth}{@{\extracolsep{\fill}}lrrrrrr@{}}
\toprule
Model & \multicolumn{3}{c}{\textbf{Score (\%)}} & \multicolumn{3}{c}{\textbf{Tokens (k / task)}} \\
\cmidrule(lr){2-4}\cmidrule(lr){5-7}
 & Native & CLI ($\Delta$) & MCP ($\Delta$) & Native & CLI ($\Delta$\%) & MCP ($\Delta$\%) \\
\midrule
\addlinespace[2pt]\rowcolor{MVGroup}\multicolumn{7}{l}{\bfseries ALFWorld\strut}\\
Qwen3-4B & 50.0 & \ensuremath{44.8_{\textcolor{SVNativeUnfavorable}{-5.2}}} & \ensuremath{41.8_{\textcolor{SVNativeUnfavorable}{-8.2}}} & 295.5 & \ensuremath{384.2_{\textcolor{SVNativeUnfavorable}{+30.0\%}}} & \ensuremath{387.1_{\textcolor{SVNativeUnfavorable}{+31.0\%}}} \\
Qwen3-8B & 31.3 & \ensuremath{44.8_{\textcolor{SVNativeFavorable}{+13.4}}} & \ensuremath{35.1_{\textcolor{SVNativeFavorable}{+3.7}}} & 273.5 & \ensuremath{292.4_{\textcolor{SVNativeUnfavorable}{+6.9\%}}} & \ensuremath{356.5_{\textcolor{SVNativeUnfavorable}{+30.3\%}}} \\
Qwen3-14B & 62.7 & \ensuremath{56.7_{\textcolor{SVNativeUnfavorable}{-6.0}}} & \ensuremath{20.1_{\textcolor{SVNativeUnfavorable}{-42.5}}} & 176.9 & \ensuremath{241.9_{\textcolor{SVNativeUnfavorable}{+36.7\%}}} & \ensuremath{296.8_{\textcolor{SVNativeUnfavorable}{+67.8\%}}} \\
Qwen3-32B & 85.1 & \ensuremath{73.9_{\textcolor{SVNativeUnfavorable}{-11.2}}} & \ensuremath{78.4_{\textcolor{SVNativeUnfavorable}{-6.7}}} & 72.4 & \ensuremath{114.7_{\textcolor{SVNativeUnfavorable}{+58.3\%}}} & \ensuremath{106.5_{\textcolor{SVNativeUnfavorable}{+46.9\%}}} \\
DeepSeek-v4-Flash & 97.0 & \ensuremath{99.3_{\textcolor{SVNativeFavorable}{+2.2}}} & \ensuremath{95.5_{\textcolor{SVNativeUnfavorable}{-1.5}}} & 50.5 & \ensuremath{58.0_{\textcolor{SVNativeUnfavorable}{+14.8\%}}} & \ensuremath{56.7_{\textcolor{SVNativeUnfavorable}{+12.4\%}}} \\
Gemini-2.5-Flash & 67.2 & \ensuremath{56.7_{\textcolor{SVNativeUnfavorable}{-10.4}}} & \ensuremath{67.2_{\textcolor{SVNativeZero}{0.0}}} & 86.7 & \ensuremath{120.0_{\textcolor{SVNativeUnfavorable}{+38.3\%}}} & \ensuremath{113.8_{\textcolor{SVNativeUnfavorable}{+31.2\%}}} \\
\addlinespace[2pt]\rowcolor{MVGroup}\multicolumn{7}{l}{\bfseries WebShop\strut}\\
Qwen3-4B & 62.9 & \ensuremath{58.3_{\textcolor{SVNativeUnfavorable}{-4.6}}} & \ensuremath{64.1_{\textcolor{SVNativeFavorable}{+1.2}}} & 32.5 & \ensuremath{37.6_{\textcolor{SVNativeUnfavorable}{+15.9\%}}} & \ensuremath{36.3_{\textcolor{SVNativeUnfavorable}{+11.7\%}}} \\
Qwen3-8B & 57.7 & \ensuremath{59.5_{\textcolor{SVNativeFavorable}{+1.8}}} & \ensuremath{56.8_{\textcolor{SVNativeUnfavorable}{-0.9}}} & 66.2 & \ensuremath{89.6_{\textcolor{SVNativeUnfavorable}{+35.4\%}}} & \ensuremath{78.8_{\textcolor{SVNativeUnfavorable}{+19.0\%}}} \\
Qwen3-14B & 54.9 & \ensuremath{59.5_{\textcolor{SVNativeFavorable}{+4.6}}} & \ensuremath{64.4_{\textcolor{SVNativeFavorable}{+9.5}}} & 92.9 & \ensuremath{83.1_{\textcolor{SVNativeFavorable}{-10.5\%}}} & \ensuremath{48.1_{\textcolor{SVNativeFavorable}{-48.3\%}}} \\
Qwen3-32B & 64.0 & \ensuremath{61.6_{\textcolor{SVNativeUnfavorable}{-2.4}}} & \ensuremath{65.7_{\textcolor{SVNativeFavorable}{+1.7}}} & 85.8 & \ensuremath{80.4_{\textcolor{SVNativeFavorable}{-6.3\%}}} & \ensuremath{88.6_{\textcolor{SVNativeUnfavorable}{+3.3\%}}} \\
DeepSeek-v4-Flash & 86.0 & \ensuremath{83.0_{\textcolor{SVNativeUnfavorable}{-3.0}}} & \ensuremath{83.5_{\textcolor{SVNativeUnfavorable}{-2.5}}} & 63.8 & \ensuremath{69.1_{\textcolor{SVNativeUnfavorable}{+8.3\%}}} & \ensuremath{65.1_{\textcolor{SVNativeUnfavorable}{+2.1\%}}} \\
Gemini-2.5-Flash & 82.2 & \ensuremath{84.5_{\textcolor{SVNativeFavorable}{+2.3}}} & \ensuremath{83.3_{\textcolor{SVNativeFavorable}{+1.0}}} & 54.9 & \ensuremath{42.8_{\textcolor{SVNativeFavorable}{-22.1\%}}} & \ensuremath{41.9_{\textcolor{SVNativeFavorable}{-23.8\%}}} \\
\addlinespace[2pt]\rowcolor{MVGroup}\multicolumn{7}{l}{\bfseries ScienceWorld\strut}\\
Qwen3-4B & 37.5 & \ensuremath{41.0_{\textcolor{SVNativeFavorable}{+3.5}}} & \ensuremath{37.9_{\textcolor{SVNativeFavorable}{+0.4}}} & 100.8 & \ensuremath{106.9_{\textcolor{SVNativeUnfavorable}{+6.1\%}}} & \ensuremath{108.7_{\textcolor{SVNativeUnfavorable}{+7.9\%}}} \\
Qwen3-8B & 51.1 & \ensuremath{46.9_{\textcolor{SVNativeUnfavorable}{-4.2}}} & \ensuremath{53.0_{\textcolor{SVNativeFavorable}{+1.9}}} & 82.6 & \ensuremath{82.5_{\textcolor{SVNativeFavorable}{-0.2\%}}} & \ensuremath{101.1_{\textcolor{SVNativeUnfavorable}{+22.5\%}}} \\
Qwen3-14B & 35.8 & \ensuremath{39.5_{\textcolor{SVNativeFavorable}{+3.7}}} & \ensuremath{35.1_{\textcolor{SVNativeUnfavorable}{-0.7}}} & 60.6 & \ensuremath{73.4_{\textcolor{SVNativeUnfavorable}{+21.0\%}}} & \ensuremath{76.0_{\textcolor{SVNativeUnfavorable}{+25.3\%}}} \\
Qwen3-32B & 63.9 & \ensuremath{61.3_{\textcolor{SVNativeUnfavorable}{-2.6}}} & \ensuremath{60.7_{\textcolor{SVNativeUnfavorable}{-3.2}}} & 73.7 & \ensuremath{80.9_{\textcolor{SVNativeUnfavorable}{+9.9\%}}} & \ensuremath{83.8_{\textcolor{SVNativeUnfavorable}{+13.8\%}}} \\
DeepSeek-v4-Flash & 81.3 & \ensuremath{77.5_{\textcolor{SVNativeUnfavorable}{-3.7}}} & \ensuremath{75.3_{\textcolor{SVNativeUnfavorable}{-6.0}}} & 77.2 & \ensuremath{97.4_{\textcolor{SVNativeUnfavorable}{+26.1\%}}} & \ensuremath{93.1_{\textcolor{SVNativeUnfavorable}{+20.5\%}}} \\
Gemini-2.5-Flash & 52.4 & \ensuremath{51.2_{\textcolor{SVNativeUnfavorable}{-1.2}}} & \ensuremath{54.0_{\textcolor{SVNativeFavorable}{+1.6}}} & 61.0 & \ensuremath{65.2_{\textcolor{SVNativeUnfavorable}{+7.0\%}}} & \ensuremath{59.2_{\textcolor{SVNativeFavorable}{-2.8\%}}} \\
\bottomrule
\end{tabular*}\par\endgroup
    \caption{
    \textbf{Task performance and interaction cost under matched interfaces.}
    MCP and CLI expose the same executable capabilities through different
    model-facing representations. Native is included as a calibration reference.
    }
    \label{tab:main_results}
\end{table}

\textbf{Matched capabilities produce substantial endpoint differences.}
Table~\ref{tab:main_results} shows that both the direction and magnitude of the
interface effect vary across model--environment pairs. For Qwen3-14B, MCP
trails CLI by 36.6 score points on ALFWorld ($20.1$ vs.\ $56.7$) but exceeds
it by 4.9 points on WebShop ($64.4$ vs.\ $59.5$). The interaction-cost effect
reverses at the same time. MCP uses 54.9k more tokens per ALFWorld task but
35.0k fewer on WebShop. Comparable reversals occur across the model suite,
showing that task performance and interaction cost respond differently to the
same interface change.

Interface representation therefore acts as a model- and environment-dependent
factor in the interaction loop, redistributing difficulty across operation
selection, argument grounding, and action realization. The following analyses
trace these endpoint differences to changes in semantic decisions, failures,
and recovery.

\subsection{Failure and Recovery}
\label{sec:rq2}

\begin{figure}[!htb]
    \centering
    \includegraphics[width=\linewidth]
    {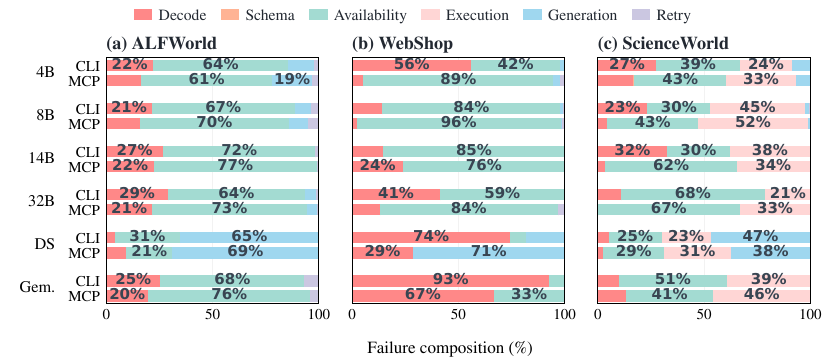}
    \caption{
    \textbf{Failure pathways under MCP and CLI.}
    Bars show the composition of failure stages among unsuccessful attempts.
    The dominant failure pathway changes with both the interface and environment.
    }
    \label{fig:failure_composition}
\end{figure}

\textbf{Interface representation changes semantic action selection after
successful decoding.}
The effect therefore extends beyond surface realization. For
WebShop/Qwen3-14B, MCP shifts decoded action frequency by $+6$ percentage
points toward \texttt{click} and $-6$ points from \texttt{search}. ALFWorld
and ScienceWorld exhibit corresponding redistributions across their larger
operation sets (Appendix~\ref{app:rq2_current_results}). The shared semantic
action space is thus exercised differently under the two interface
representations.

\textbf{Interfaces reshape both failure incidence and failure location.}
Figure~\ref{fig:failure_composition} shows that unsuccessful attempts enter
different stages of the action pipeline under MCP and CLI. On WebShop, MCP
lowers the attempt-level failure rate for all six models by 1.1--8.9
percentage points, while the remaining failures shift from the stronger
surface-decoding component under CLI toward later availability and execution
stages. ScienceWorld exhibits a different failure composition, with
availability and execution accounting for a larger share of unsuccessful
attempts. Interface representation therefore determines both the frequency
and stage of interaction failure.

\textbf{Recovery behavior carries interface effects into subsequent
decisions.}
After failure feedback, the next response can preserve the semantic decision
and repair its realization, select a different action, repeat the error, or
enter another failure mode. On WebShop/Qwen3-4B, retry acceptance is $50\%$
under CLI and $77\%$ under MCP, while later task success is $18\%$ and
$34\%$, respectively (Appendix~\ref{app:rq2_current_results}). These
interface-specific recovery paths update the visible interaction history and
thereby change the context conditioning later decisions.

\subsection{Local Effects and Trajectory Propagation}
\label{sec:rq3}

\begin{figure}[!htb]
    \centering
    \includegraphics[width=\linewidth]
    {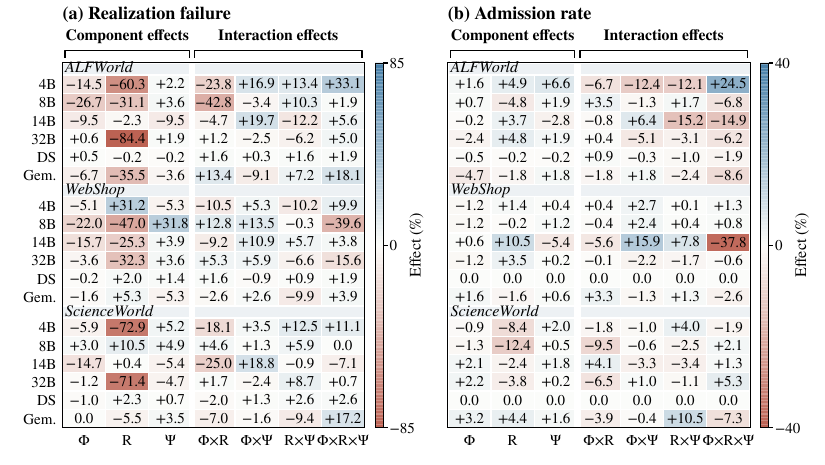}
    \caption{
    \textbf{Component and interaction effects on local response behavior.}
    Effects on (a) realization failure and (b) admission given decoding are shown
    for presentation $\Phi$, realization $R$, and feedback $\Psi$. Main effects
    compare CLI with MCP while averaging over the remaining blocks. Interaction
    terms capture non-additivity under the factorial coding. Each panel uses a
    symmetric percentage-point scale centered at zero.
    }
    \label{fig:factorial_effects}
\end{figure}

\textbf{Local interface effects are large and strongly compositional.}
Figure~\ref{fig:factorial_effects} separates presentation, realization, and
feedback at matched decision states. Realization produces the largest local
shifts in several strata. Its realization-failure effect reaches 84.4
percentage points for ALFWorld/Qwen3-32B and 72.9 points for
ScienceWorld/Qwen3-4B. Interaction terms are also substantial. For example,
the three-way interaction in WebShop/Qwen3-14B changes admission by 37.8
points. The behavioral effect of each interface component therefore depends
on the surrounding interface configuration.

\textbf{Local divergence propagates through the interaction trajectory.}
Across 2,004 paired MCP--CLI episodes, 1,889 ($94.3\%$) exhibit a structured
divergence, with first divergence occurring near the beginning of interaction
across model--environment strata
(Appendix~\ref{app:rq3_current_results}). Semantic-action differences account
for the dominant share of divergent pairs, with realization differences
providing a recurring second source. After divergence, the paired trajectories
accumulate different token costs and attempt counts and reach distinct terminal
outcomes across settings. MICA thereby connects matched-state interface
sensitivity to the trajectory-level performance and cost differences observed
in Section~\ref{sec:rq1}.

\subsection{DART Evaluation}
\label{sec:rq4}

\begin{figure}[!htb]
    \centering
    \includegraphics[width=\linewidth]
    {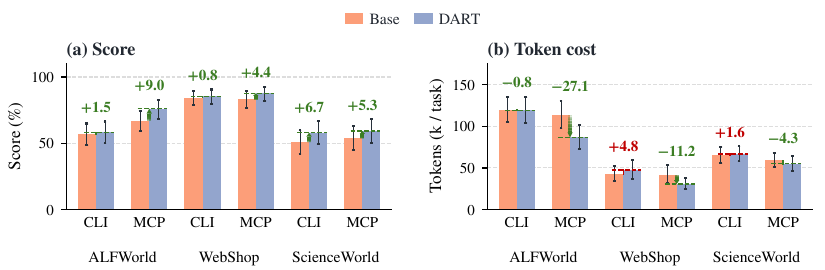}
    \caption{
    \textbf{Effect of DART on task performance and interaction cost.}
    Each Gemini-2.5-Flash Base--DART pair matches the benchmark, interface,
    task set, generation settings, and episode budget. Labels show the
    DART-minus-Base change.
    }
    \label{fig:dart_policy}
\end{figure}

\textbf{DART improves task performance in every evaluated setting.}
Figure~\ref{fig:dart_policy} reports gains of $+1.5/+9.0$ points on ALFWorld,
$+0.8/+4.4$ on WebShop, and $+6.7/+5.3$ on ScienceWorld for CLI/MCP,
respectively. Under MCP, these gains are accompanied by token reductions of
27.1k, 11.2k, and 4.3k per task across the three environments. Under CLI,
token cost changes by $-0.8$k, $+4.8$k, and $+1.6$k.

\textbf{Commitment preservation concentrates recovery at the realization
boundary.}
DART keeps the validated semantic commitment fixed and changes only its
interface-specific realization. The event-level analysis in
Appendix~\ref{app:rq4_current_results} shows that $71$--$100\%$ of triggered
interventions are admitted on the first realization, and three of the four
preserving-correction opportunities are admitted after a single correction.
The preserving branch therefore resolves interface-specific realization
failures while keeping semantic selection fixed, directly linking the mechanism
identified by MICA to the score improvements in
Figure~\ref{fig:dart_policy}.
\section{Discussion}
\label{sec:discussion}

\paragraph{Interface representation is part of the decision process.}
Under matched executable capabilities, MCP and CLI share the same semantic
action contract and downstream runtime yet induce different policies because
they present, realize, and return feedback about those capabilities differently.
Interface design should therefore be treated as part of the agent's
decision environment rather than as a passive access layer.

\paragraph{Trajectory-level behavior emerges from local interface effects.}
MICA links local sensitivity to downstream behavior by combining matched-state
probes with paired natural trajectories. Presentation, realization, and feedback
can change a local response, after which the resulting history and state
condition subsequent decisions. Later semantic realignment can occur
after the trajectories have already followed different histories, making the
trajectory the appropriate unit for analyzing interactive agent behavior.

\paragraph{Semantic commitment separates decision from realization during recovery.}
DART turns this distinction into an explicit controller boundary. Once a
semantic action is validated, recovery preserves that commitment and repairs
only its interface-specific realization. Language-agent controllers can improve
recovery by preserving established semantic decisions and restricting repair to
the representation-specific component that requires correction.

\section{Conclusion}
\label{sec:conclusion}

Our study establishes interface representation as a consequential component of
interactive language-agent behavior under matched executable capabilities.
MICA combines contract-matched interfaces, matched-state probes, and paired
natural trajectories to connect local representational sensitivity with
downstream differences in semantic decisions, recovery, interaction cost, and
task outcomes. Building on this mechanism, DART preserves validated semantic
commitments while repairing interface-specific realization and improves
benchmark score in all six evaluated benchmark--interface settings. Together,
these results show that interface design shapes the effective decision
environment through which an agent acts. For interactive language agents,
how capabilities are represented is part of how the resulting policy is
realized through interaction.
\section*{Ethics Statement}

This work operates within existing action, validation, and execution boundaries
and does not expand the executable capabilities available to the agent. MICA
varies only the model-facing representation of a shared semantic action
contract while preserving the underlying capability set and downstream runtime.
DART likewise operates within the same contract and execution pipeline by
preserving validated semantic commitments during interface-specific
realization repair. The study therefore focuses on understanding and improving
the reliability of tool-mediated agent interaction without introducing
additional action authority or altering the safeguards defined by the
evaluation environment.

\section*{Reproducibility Statement}

We provide the experimental details required to reproduce
the study throughout the main paper and appendices. Section~\ref{sec:setting} defines the
semantic comparison boundary, interface factorization, and consequential
divergence. Section~\ref{sec:framework} specifies MICA and DART, while Section~\ref{sec:experiments} summarizes the
controlled evaluation setting. Appendix~\ref{app:evaluation_protocol} provides the task sets, generation
settings, interaction budgets, evaluation protocols, metrics, and statistical
analysis. Appendix~\ref{app:interfaces} specifies the contract-matched MCP and CLI construction
and conformance checks. Appendix~\ref{app:dart_implementation} gives the DART controller, trigger
configuration, and evaluation protocol. Appendix~\ref{app:prompts} provides the ordinary
policy and phase-specific response contracts used in the experiments. \CodeAvailabilityStatement

\section*{AI Use Statement}

Generative AI tools were used as assistive tools during the preparation of this
work, including language polishing, code implementation assistance, and the
preparation of figures.
All AI-assisted material was reviewed and incorporated by the authors, who
retain full responsibility for the technical content and final presentation
of the paper.

\FloatBarrier

\clearpage
\appendix
\setcounter{figure}{0}
\setcounter{table}{0}
\renewcommand{\thefigure}{S\arabic{figure}}
\renewcommand{\thetable}{S\arabic{table}}
\renewcommand{\theHfigure}{S\arabic{figure}}
\renewcommand{\theHtable}{S\arabic{table}}
\begingroup
\makeatletter
\newcommand{\AppTableCaptionStyle}{%
  \long\def\@makecaption##1##2{%
    \vskip 3pt\noindent
    {\normalsize\raggedright ##1: ##2\par}\vskip 5pt}}
\newcommand{\AppPreferHere}{%
  \let\AppOriginalXFloat\@xfloat
  \def\@xfloat##1[##2]{\AppOriginalXFloat{##1}[!htbp]}}
\makeatother

\definecolor{AppFrameGray}{gray}{.50}
\definecolor{AppBodyGray}{gray}{.976}
\newcolumntype{J}{>{\raggedright\arraybackslash}X}
\newenvironment{AppTable}[1][!htbp]{%
  \begin{table}[#1]\centering\AppTableCaptionStyle
  \fontsize{9}{10.8}\selectfont\setlength{\tabcolsep}{4pt}%
  \renewcommand{\arraystretch}{1.10}%
  \setlength{\heavyrulewidth}{.7pt}\setlength{\lightrulewidth}{.4pt}%
}{\end{table}}
\newcommand{\AppTableNote}[1]{\par\vspace{4pt}%
  {\fontsize{9}{10.8}\selectfont\raggedright\noindent #1\par}}
\newcommand{\AppLiteral}[1]{{\fontencoding{T1}\upshape\ttfamily\selectfont #1}}
\newcommand{\AppCode}[1]{\par\begingroup
  \fontencoding{T1}\ttfamily\upshape\fontsize{9.5}{12.5}\selectfont
  \setlength{\parskip}{0pt}\setlength{\parindent}{0pt}\raggedright
  #1\par\endgroup}

\ExplSyntaxOn
\seq_new:N \l_app_case_rows_seq
\seq_new:N \l_app_case_paragraphs_seq
\tl_new:N \l_app_case_output_tl
\tl_new:N \l_app_case_label_tl
\cs_new_protected:Npn \app_case_row:w \rowcolor #1 #2 & #3 \q_stop
  {
    \seq_set_split:Nnn \l_app_case_paragraphs_seq {\par} {#3}
    \tl_set:Nn \l_app_case_label_tl {#2}
    \seq_map_inline:Nn \l_app_case_paragraphs_seq
      {
        \tl_if_blank:nF {##1}
          {
            \tl_put_right:Nx \l_app_case_output_tl
              {
                \exp_not:N \rowcolor {\exp_not:n {#1}}
                \exp_not:V \l_app_case_label_tl & \exp_not:n {##1}
                \exp_not:N \\
              }
            \tl_clear:N \l_app_case_label_tl
          }
      }
    \tl_put_right:Nn \l_app_case_output_tl {\noalign{\vskip 2pt}}
  }
\NewDocumentCommand{\DARTPrepareRows}{+m}
  {
    \tl_clear:N \l_app_case_output_tl
    \seq_set_split:Nnn \l_app_case_rows_seq {\\[2pt]} {#1}
    \seq_map_inline:Nn \l_app_case_rows_seq
      {
        \tl_set:Nn \l_tmpa_tl {##1}
        \tl_trim_spaces:N \l_tmpa_tl
        \regex_replace_once:nnN {\A(\s|\c{par})+} {} \l_tmpa_tl
        \tl_if_empty:NF \l_tmpa_tl
          {\exp_after:wN \app_case_row:w \l_tmpa_tl \q_stop}
      }
  }
\NewExpandableDocumentCommand{\DARTPreparedRows}{} {\tl_use:N \l_app_case_output_tl}
\ExplSyntaxOff

\newcounter{appprompt}
\newcounter{appexample}
\newcounter{appalgorithm}
\newcounter{appmapping}
\renewcommand{\theappprompt}{S\arabic{appprompt}}
\renewcommand{\theappexample}{S\arabic{appexample}}
\renewcommand{\theappalgorithm}{S\arabic{appalgorithm}}
\providecommand{\theHappprompt}{}
\providecommand{\theHappexample}{}
\providecommand{\theHappalgorithm}{}
\renewcommand{\theHappprompt}{app.prompt.\arabic{appprompt}}
\renewcommand{\theHappexample}{app.example.\arabic{appexample}}
\renewcommand{\theHappalgorithm}{app.algorithm.\arabic{appalgorithm}}

\newcommand{\AppFrame}[5]{\par
  \Needspace{3\baselineskip}%
  \refstepcounter{#1}\label{#2}%
  \begin{tcolorbox}[
    enhanced,breakable,arc=2pt,boxrule=.7pt,
    colframe=AppFrameGray,colback=AppBodyGray,colbacktitle=AppFrameGray,
    coltitle=white,fonttitle=\fontsize{10}{11.5}\selectfont\bfseries,
    title={#3\csname the#1\endcsname.\enspace #4},
    title after break={#3\csname the#1\endcsname.\enspace #4\ (continued)},
    left=12pt,right=12pt,top=6pt,bottom=6pt,boxsep=0pt,
    toptitle=2.4pt,bottomtitle=2.4pt,before skip=8pt,after skip=6pt,
    pad at break*=4pt,lines before break=3,
    fontupper=\fontencoding{T1}\fontsize{10}{12.5}\selectfont]
    \setlength{\parindent}{0pt}\setlength{\parskip}{4pt}%
    \setlength{\emergencystretch}{1em}%
    #5\par
  \end{tcolorbox}}
\newcommand{\AppPrompt}[3]{\AppFrame{appprompt}{#1}{Prompt~}{#2}{\itshape #3}}
\newcommand{\AppExample}[3]{\AppFrame{appexample}{#1}{Example~}{#2}{#3}}

\newcounter{AppAlgLine}
\newenvironment{AppAlgorithm}[2]{\par\addvspace{8pt}\noindent
  \begin{minipage}{\linewidth}\small\setlength{\parskip}{0pt}
  \refstepcounter{appalgorithm}\label{#1}\setcounter{AppAlgLine}{0}
  \hrule height .7pt\vspace{4pt}\noindent
  \textbf{Algorithm \theappalgorithm}\enspace #2\par
  \vspace{4pt}\hrule height .4pt\vspace{5pt}
}{\par\vspace{4pt}\hrule height .7pt\end{minipage}\par\addvspace{7pt}}
\newcommand{\AppIO}[2]{\noindent\textbf{Input:} #1\par\vspace{2pt}
  \noindent\textbf{Output:} #2\par\vspace{5pt}}
\newcommand{\AppLine}[2]{\stepcounter{AppAlgLine}\noindent
  \makebox[1.55em][r]{\footnotesize\theAppAlgLine:}\hspace{.65em}\hspace*{#1em}%
  \parbox[t]{\dimexpr\linewidth-2.2em-#1em\relax}{\raggedright #2\strut}\par\vspace{.6pt}}
\setlength{\emergencystretch}{1em}
\makeatletter
\let\AppOriginalAddToContents\addtocontents
\def\AppMainContentsExtension{toc}
\renewcommand{\addtocontents}[2]{%
  \def\AppContentsExtension{#1}%
  \ifx\AppContentsExtension\AppMainContentsExtension
    \AppOriginalAddToContents{apptoc}{#2}%
  \else
    \AppOriginalAddToContents{#1}{#2}%
  \fi}

\begingroup
\hypersetup{linktoc=section}
\newcommand{\AppContentsSubsection}{\@dottedtocline{2}{1.5em}{2.3em}}
\renewcommand{\l@section}[2]{%
  \addvspace{1em plus 1pt}
  {\bfseries\@dottedtocline{1}{0pt}{1.5em}{\textcolor{black}{#1}}{#2}}}
\renewcommand{\l@subsection}[2]{%
  \AppContentsSubsection{\textcolor{black}{#1}}{#2}}
\section*{Appendix}
\pdfbookmark[0]{Appendix}{appendix.contents}
\c@tocdepth=2\relax
\normalsize
\setlength{\parskip}{0pt}
\@starttoc{apptoc}
\endgroup
\makeatother
\clearpage
\begingroup
\setlength{\parskip}{2pt}
\setlength{\floatsep}{8pt plus 2pt minus 2pt}
\setlength{\textfloatsep}{8pt plus 2pt minus 2pt}
\setlength{\intextsep}{6pt plus 2pt minus 2pt}
\renewcommand{\bottomfraction}{.95}
\renewcommand{\floatpagefraction}{.85}
\makeatletter
\c@topnumber=4
\c@bottomnumber=4
\c@totalnumber=6
\setlength{\@fptop}{0pt}
\setlength{\@fpsep}{8pt plus 2pt minus 2pt}
\setlength{\@fpbot}{0pt plus 1fil}
\long\def\@makecaption#1#2{\vskip 2pt\noindent
  {\fontsize{9}{10.8}\selectfont\raggedright\textbf{#1.} #2\par}\vskip 3pt}
\newcommand{\SVTable}[3]{\begin{table}[!htb]\centering
  \def\@captype{table}\caption{#3}\label{#2}
  #1
  \end{table}}
\newcommand{\SVFigureContent}[3]{%
  #1
  \def\@captype{figure}\caption{#3}\label{#2}}
\newcommand{\SVFigure}[3]{\begin{figure}[!htb]\centering
  \SVFigureContent{#1}{#2}{#3}\end{figure}}
\makeatother
\newcommand{\ResultLead}[1]{\par\noindent\textbf{#1}\ }

\section{Supplementary Results by Research Question}
\label{app:measured_artifacts}
This appendix extends the main-text results with finer-grained evidence for the
four research questions. RQ1 resolves endpoint performance and interaction cost
to individual tasks and resource components. RQ2 follows the interaction
pipeline from semantic action selection through failure and recovery. RQ3
localizes interface sensitivity at matched decision states and traces how the
first consequential divergence propagates through natural trajectories. RQ4
then examines how DART acts at the realization boundary identified by this
analysis.

\subsection{RQ1: Endpoint Performance and Interaction Cost}
\label{app:rq1_current_results}

\ResultLead{Interface representation reshapes token use at the task level.}
Figure~\ref{fig:supp_paired_token_effect} resolves the aggregate token costs in
Table~\ref{tab:main_results} into paired task-level effects. The distributions
show substantial within-stratum variation. MCP uses more tokens on
some tasks and fewer on others even for the same model and environment. The
aggregate cost differences in Table~\ref{tab:main_results} therefore arise from
systematic shifts in the distribution of task-level interaction rather than from
a uniform interface overhead.

\SVFigure{\includegraphics[width=\linewidth]{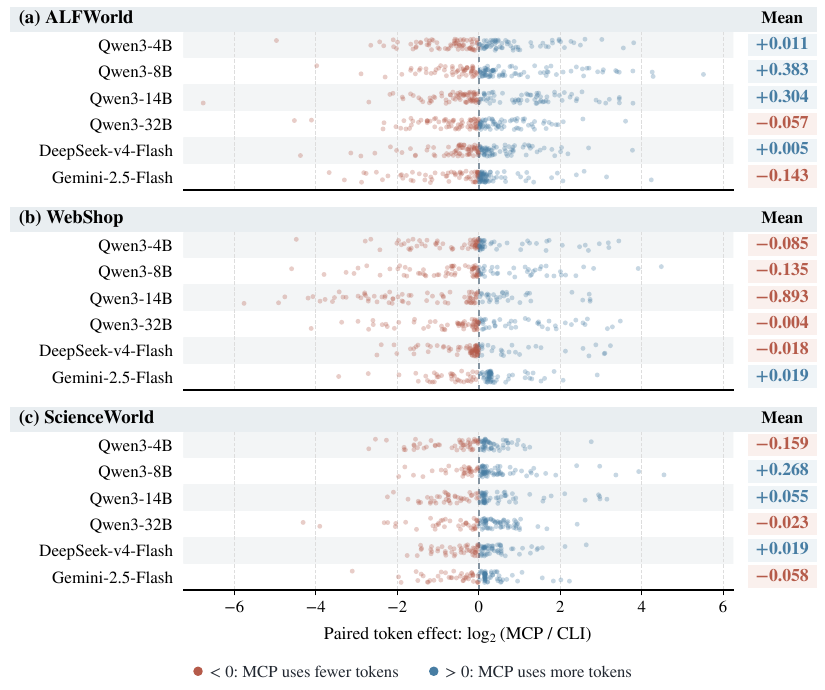}}{fig:supp_paired_token_effect}{Paired task-level
token effects for the CLI/MCP configurations in Tables~\ref{tab:main_results}
and~\ref{tab:secondary_effects}. Each point reports
$\log_2(T_{\mathrm{MCP}}/T_{\mathrm{CLI}})$ for one task after within-task
replicate averaging. Negative values indicate lower token use under MCP and
positive values indicate higher token use. The right column reports the
model-level mean within each environment.}

\ResultLead{Interaction cost is shaped jointly by call frequency and token use per call.}
Table~\ref{tab:secondary_effects} decomposes total token use into the number of
model calls and the tokens consumed per call, revealing distinct mechanisms
behind the endpoint cost shifts. On ALFWorld, Qwen3-14B under MCP increases
from $46.9$ to $57.1$ calls per task while tokens per call remain nearly
unchanged ($4.46$k versus $4.50$k), showing that its higher cost is driven
primarily by a longer interaction trajectory. On WebShop, the same model shows
the opposite pattern. MCP reduces both calls per task ($14.2 \rightarrow 9.6$)
and tokens per call ($4.26$k $\rightarrow 3.23$k), producing a substantially
lighter interaction profile. These results show that interface representation
changes not only how much computation an agent consumes, but also how that
computation is distributed across the trajectory.

\SVTable{\begingroup\centering\fontsize{8}{9.6}\selectfont
\setlength{\tabcolsep}{2pt}\renewcommand{\arraystretch}{1.02}
\setlength{\aboverulesep}{2pt}\setlength{\belowrulesep}{2pt}
\begin{tabular*}{\linewidth}{@{\extracolsep{\fill}}lrrrrrr@{}}
\toprule
Model & \multicolumn{2}{c}{\textbf{Success (\%)}} & \multicolumn{2}{c}{\textbf{Calls / task}} & \multicolumn{2}{c}{\textbf{Tokens / call (k)}} \\
\cmidrule(lr){2-3}\cmidrule(lr){4-5}\cmidrule(lr){6-7}
 & CLI & MCP ($\Delta$) & CLI & MCP ($\Delta$) & CLI & MCP ($\Delta$) \\
\midrule
\addlinespace[2pt]\rowcolor{MVGroup}\multicolumn{7}{l}{\bfseries ALFWorld\strut}\\
Qwen3-4B & 44.8 & 41.8\,\SVDelta{SVNegative}{-3.0} & 53.1 & 51.5\,\SVDelta{SVNegative}{-1.6} & 5.96 & 6.23\,\SVDelta{SVPositive}{+0.27} \\
Qwen3-8B & 44.8 & 35.1\,\SVDelta{SVNegative}{-9.7} & 47.3 & 53.2\,\SVDelta{SVPositive}{+5.9} & 5.34 & 5.89\,\SVDelta{SVPositive}{+0.56} \\
Qwen3-14B & 56.7 & 20.1\,\SVDelta{SVNegative}{-36.6} & 46.9 & 57.1\,\SVDelta{SVPositive}{+10.2} & 4.46 & 4.50\,\SVDelta{SVPositive}{+0.04} \\
Qwen3-32B & 73.9 & 78.4\,\SVDelta{SVPositive}{+4.5} & 27.8 & 26.6\,\SVDelta{SVNegative}{-1.2} & 3.54 & 3.50\,\SVDelta{SVNegative}{-0.04} \\
DeepSeek-v4-Flash & 99.3 & 95.5\,\SVDelta{SVNegative}{-3.7} & 15.2 & 15.3\,\SVDelta{SVPositive}{+0.1} & 3.10 & 3.08\,\SVDelta{SVNegative}{-0.02} \\
Gemini-2.5-Flash & 56.7 & 67.2\,\SVDelta{SVPositive}{+10.4} & 30.2 & 27.1\,\SVDelta{SVNegative}{-3.1} & 3.48 & 3.55\,\SVDelta{SVPositive}{+0.07} \\
\addlinespace[2pt]\rowcolor{MVGroup}\multicolumn{7}{l}{\bfseries WebShop\strut}\\
Qwen3-4B & 17.0 & 22.0\,\SVDelta{SVPositive}{+5.0} & 8.6 & 7.9\,\SVDelta{SVNegative}{-0.7} & 3.49 & 3.48\,\SVDelta{SVNegative}{-0.01} \\
Qwen3-8B & 23.0 & 21.0\,\SVDelta{SVNegative}{-2.0} & 14.4 & 12.7\,\SVDelta{SVNegative}{-1.7} & 4.48 & 4.48\,\SVDelta{SVNegative}{-0.01} \\
Qwen3-14B & 22.0 & 35.0\,\SVDelta{SVPositive}{+13.0} & 14.2 & 9.6\,\SVDelta{SVNegative}{-4.6} & 4.26 & 3.23\,\SVDelta{SVNegative}{-1.03} \\
Qwen3-32B & 35.0 & 39.0\,\SVDelta{SVPositive}{+4.0} & 13.8 & 14.0\,\SVDelta{SVPositive}{+0.2} & 4.33 & 4.53\,\SVDelta{SVPositive}{+0.19} \\
DeepSeek-v4-Flash & 79.0 & 82.0\,\SVDelta{SVPositive}{+3.0} & 11.2 & 10.9\,\SVDelta{SVNegative}{-0.3} & 4.13 & 4.10\,\SVDelta{SVNegative}{-0.03} \\
Gemini-2.5-Flash & 77.0 & 74.0\,\SVDelta{SVNegative}{-3.0} & 9.1 & 8.4\,\SVDelta{SVNegative}{-0.7} & 3.57 & 3.92\,\SVDelta{SVPositive}{+0.35} \\
\addlinespace[2pt]\rowcolor{MVGroup}\multicolumn{7}{l}{\bfseries ScienceWorld\strut}\\
Qwen3-4B & 26.0 & 24.0\,\SVDelta{SVNegative}{-2.0} & 17.0 & 16.3\,\SVDelta{SVNegative}{-0.8} & 6.00 & 6.21\,\SVDelta{SVPositive}{+0.21} \\
Qwen3-8B & 35.0 & 37.0\,\SVDelta{SVPositive}{+2.0} & 14.0 & 15.8\,\SVDelta{SVPositive}{+1.9} & 5.48 & 6.01\,\SVDelta{SVPositive}{+0.53} \\
Qwen3-14B & 28.0 & 24.0\,\SVDelta{SVNegative}{-4.0} & 14.0 & 13.8\,\SVDelta{SVNegative}{-0.2} & 4.78 & 5.06\,\SVDelta{SVPositive}{+0.28} \\
Qwen3-32B & 46.0 & 47.0\,\SVDelta{SVPositive}{+1.0} & 14.7 & 13.9\,\SVDelta{SVNegative}{-0.8} & 5.31 & 5.76\,\SVDelta{SVPositive}{+0.44} \\
DeepSeek-v4-Flash & 64.0 & 66.0\,\SVDelta{SVPositive}{+2.0} & 13.6 & 12.7\,\SVDelta{SVNegative}{-0.8} & 6.51 & 6.74\,\SVDelta{SVPositive}{+0.23} \\
Gemini-2.5-Flash & 42.0 & 48.0\,\SVDelta{SVPositive}{+6.0} & 10.5 & 9.1\,\SVDelta{SVNegative}{-1.4} & 5.87 & 6.31\,\SVDelta{SVPositive}{+0.44} \\
\bottomrule
\end{tabular*}\par\endgroup}{tab:secondary_effects}{Secondary endpoint and resource
statistics for the CLI/MCP settings in Table~\ref{tab:main_results}. We report
task success, model calls per task, and tokens per call. Parenthesized values give
the MCP-minus-CLI difference for each metric.}

\subsection{RQ2: Semantic Decisions, Failures, and Recovery}
\label{app:rq2_current_results}
\label{app:recovery_analysis}

\ResultLead{Interface representation changes semantic action selection beyond surface decoding.}
Figure~\ref{fig:supp_semantic} compares operation frequencies among responses
that decode into typed proposals, isolating semantic choice from surface-format
failure. The resulting shifts are structured but model- and environment-dependent.
On WebShop, for example, Qwen3-14B moves by $+6$ percentage points toward
\texttt{click} and $-6$ points from \texttt{search} under MCP. ALFWorld and
ScienceWorld show broader redistributions across navigation, inspection,
manipulation, and other operations.

\SVFigure{\includegraphics[width=\linewidth]{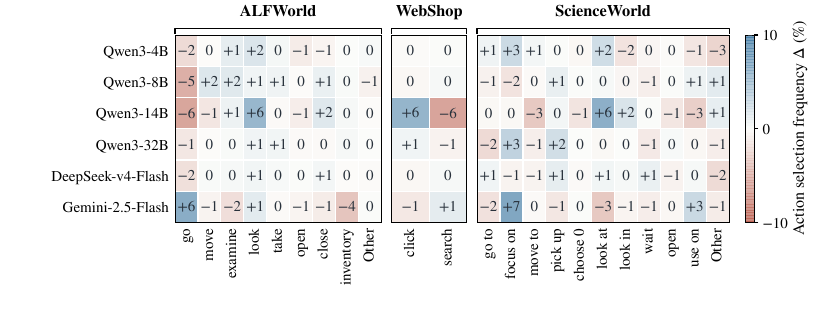}}{fig:supp_semantic}{MCP-minus-CLI changes in
semantic action-selection frequency (\%), conditional on successful decoding.
Models are shown by row and operations by environment. The Other category aggregates the
remaining operations.}

\begingroup\setlength{\parskip}{0pt}
\makeatletter
\long\def\@makecaption#1#2{\noindent
  {\fontsize{9}{10.8}\selectfont\raggedright\textbf{#1.} #2\par}\vskip 2pt}
\makeatother

\ResultLead{Interfaces reshape both failure frequency and failure pathway.}
Table~\ref{tab:failure_incidence} complements Figure~\ref{fig:failure_composition}
by measuring how often attempts fail, while the main-text figure resolves where
those failures occur in the action pipeline. WebShop provides a clear example.
MCP has lower attempt-level failure rates for all six models, with absolute
reductions ranging from $1.1$ to $8.9$ percentage points. At the same time, the
composition of the remaining failures differs markedly across models, shifting
among decoding, availability, execution, generation, and retry stages.

\SVTable{\begingroup\centering\fontsize{9}{10.6}\selectfont
\setlength{\tabcolsep}{2pt}\renewcommand{\arraystretch}{1.0}
\setlength{\aboverulesep}{1pt}\setlength{\belowrulesep}{1pt}
\begin{tabular*}{\linewidth}{@{\extracolsep{\fill}}lrrrrrrrrr@{}}
\toprule
\raisebox{-0.5\dimexpr\ht\strutbox+\dp\strutbox+\aboverulesep+\belowrulesep+\cmidrulewidth\relax}[0pt][0pt]{Model} & \multicolumn{3}{c}{\textbf{ALFWorld}} & \multicolumn{3}{c}{\textbf{WebShop}} & \multicolumn{3}{c}{\textbf{ScienceWorld}} \\
\cmidrule(lr){2-4}\cmidrule(lr){5-7}\cmidrule(lr){8-10}
 & CLI & MCP & $\Delta$ & CLI & MCP & $\Delta$ & CLI & MCP & $\Delta$ \\
\midrule
Qwen3-4B & 31.4 & 29.6 & \textcolor{SVNegative}{\ensuremath{-1.9}} & 20.8 & 12.1 & \textcolor{SVNegative}{\ensuremath{-8.8}} & 22.3 & 23.8 & \textcolor{SVPositive}{\ensuremath{+1.5}} \\
Qwen3-8B & 20.3 & 23.2 & \textcolor{SVPositive}{\ensuremath{+2.9}} & 31.2 & 22.4 & \textcolor{SVNegative}{\ensuremath{-8.9}} & 22.2 & 20.2 & \textcolor{SVNegative}{\ensuremath{-2.0}} \\
Qwen3-14B & 22.6 & 22.6 & \textcolor{SVZero}{\ensuremath{0.0}} & 12.0 & 10.8 & \textcolor{SVNegative}{\ensuremath{-1.1}} & 20.9 & 20.7 & \textcolor{SVNegative}{\ensuremath{-0.2}} \\
Qwen3-32B & 5.7 & 4.6 & \textcolor{SVNegative}{\ensuremath{-1.2}} & 14.2 & 10.9 & \textcolor{SVNegative}{\ensuremath{-3.2}} & 6.4 & 7.7 & \textcolor{SVPositive}{\ensuremath{+1.4}} \\
DeepSeek-v4-Flash & 3.7 & 2.1 & \textcolor{SVNegative}{\ensuremath{-1.6}} & 3.5 & 1.3 & \textcolor{SVNegative}{\ensuremath{-2.2}} & 7.4 & 6.3 & \textcolor{SVNegative}{\ensuremath{-1.1}} \\
Gemini-2.5-Flash & 1.5 & 2.1 & \textcolor{SVPositive}{\ensuremath{+0.6}} & 1.5 & 0.4 & \textcolor{SVNegative}{\ensuremath{-1.2}} & 10.1 & 9.1 & \textcolor{SVNegative}{\ensuremath{-1.1}} \\
\bottomrule
\end{tabular*}\par\endgroup}{tab:failure_incidence}{Attempt-level failure rates
(\%) under CLI and MCP, including retry attempts. $\Delta$ reports the
MCP-minus-CLI difference.}

\SVFigure{\includegraphics[width=\linewidth]{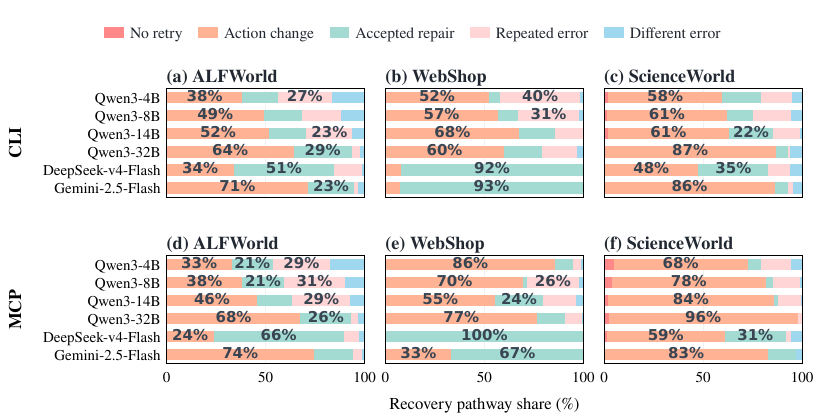}}{fig:supp_recovery}{Recovery pathways following eligible
failures under CLI and MCP. Bars report the share of failures followed by no
retry, semantic action change, accepted repair, repeated error, or a different
error.}
\SVFigure{\includegraphics[width=\linewidth]{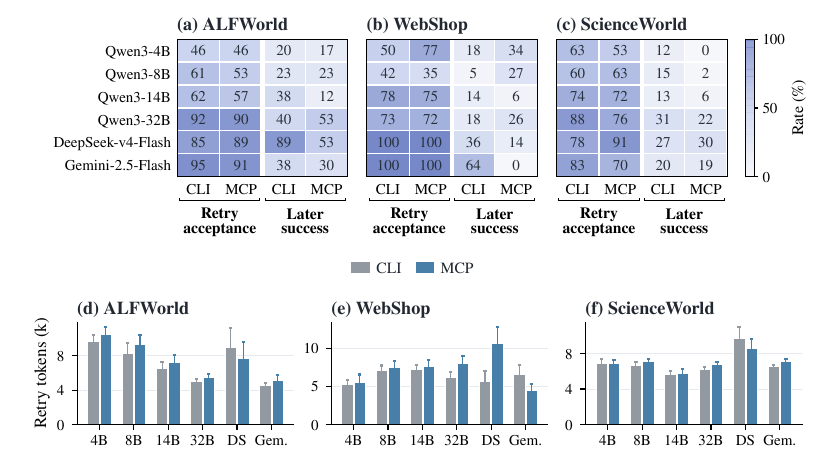}}{fig:supp_recovery_cost}{Immediate retry
acceptance, later task success, and mean retry token cost following eligible
failures. Whiskers denote 95\% task-cluster bootstrap intervals.}

\ResultLead{Recovery dynamics separate immediate correction from downstream task progress.}
Figures~\ref{fig:supp_recovery} and~\ref{fig:supp_recovery_cost} trace what
happens after an eligible failure. The next response can preserve the decision
and repair its realization, change the semantic action, repeat the error, or
move into a different failure mode. On WebShop/Qwen3-4B, for example, retry acceptance is
$50\%$ under CLI and $77\%$ under MCP, while later task success is $18\%$ and
$34\%$, respectively. Across settings, the two rates move on different scales,
showing that recovery unfolds in stages, first restoring an admissible action,
then converting that local repair into useful trajectory progress.

\endgroup

\subsection{RQ3: Local Origins and Downstream Propagation}
\label{app:rq3_current_results}

\SVFigure{\includegraphics[width=\linewidth]{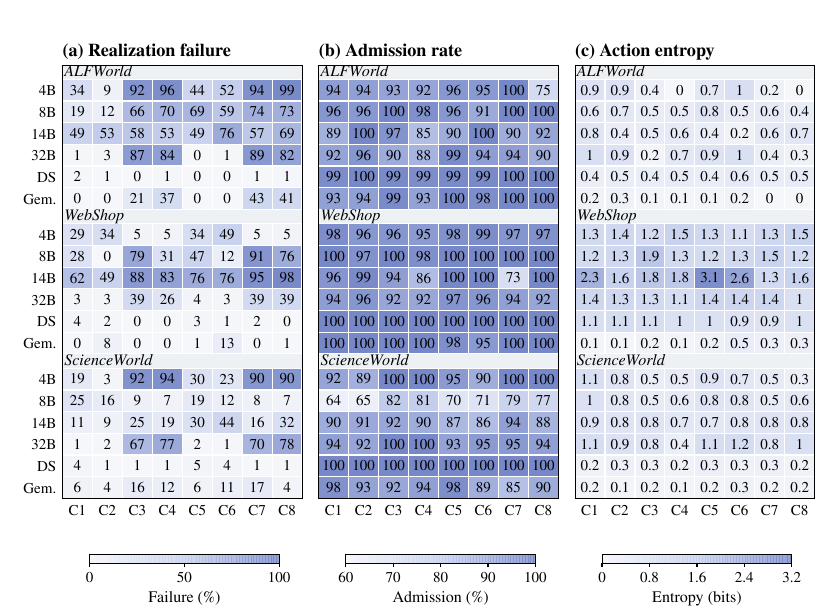}}{fig:supp_local_diagnostics}{Matched-state
response profiles for the eight combinations of presentation $\Phi$, realization
$R$, and feedback $\Psi$. Panels report realization failure, admission after
decoding, and semantic-action entropy. C1--C8 correspond to MMM, MMC, MCM, MCC,
CMM, CMC, CCM, and CCC, where M/C denotes MCP/CLI.}

\ResultLead{Interface components interact strongly at matched decision states.}
The main-text factorial analysis summarizes the component and interaction
effects, while Figure~\ref{fig:supp_local_diagnostics} exposes the eight
matched-state response profiles from which those contrasts arise. Realization is
especially sensitive to its surrounding interface context. For Qwen3-32B on
ALFWorld, switching only the realization protocol from MCP to CLI changes
realization failure from $1\%$ to $87\%$ under MCP presentation and from $0\%$
to $89\%$ under CLI presentation. Across models and environments, admission and
semantic-action entropy shift alongside realization failure, showing that
presentation, realization, and feedback jointly shape both whether an action is
successfully expressed and how decoded semantic actions are distributed.
The behavioral consequence of one interface
component depends on the configuration around it.

\begin{figure}[!htb]\centering
\SVFigureContent{\includegraphics[width=\linewidth]{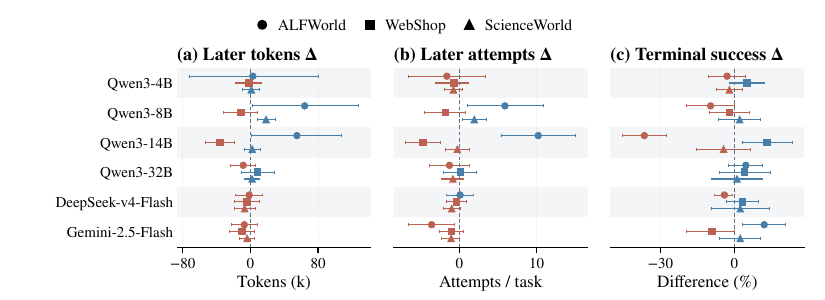}}{fig:supp_full_propagation}{Post-divergence
MCP-minus-CLI differences in charged tokens, interaction attempts, and terminal
success. Shapes identify environments and whiskers denote 95\% paired-task
bootstrap intervals.}
\end{figure}

\begin{figure}[!htb]\centering
\SVFigureContent{\includegraphics[width=\linewidth]{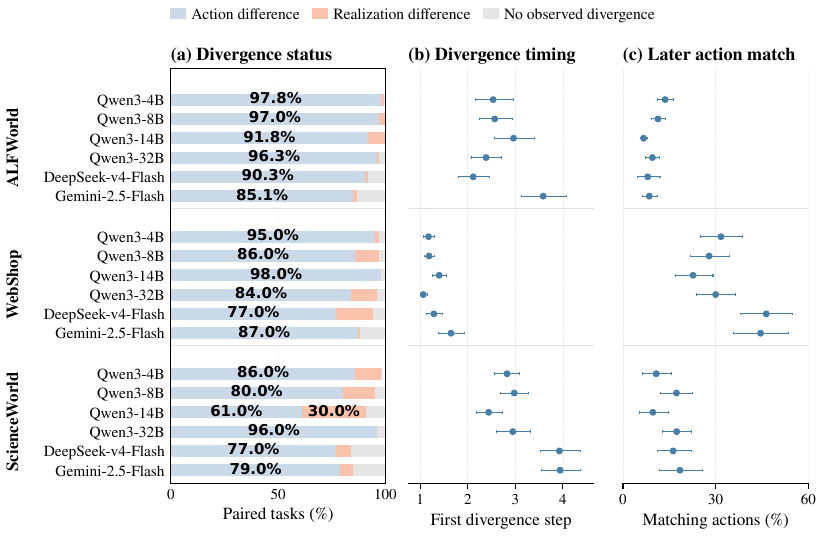}}{fig:additional_first_divergence}{Structure of
natural MCP--CLI divergence across 2,004 paired trajectories. Panels show (a) divergence type,
(b) mean first-divergence step, and (c) later semantic-action agreement. Whiskers
denote 95\% task-cluster bootstrap intervals.}
\end{figure}

\ResultLead{Consequential divergence is widespread and emerges early.}
Across the 2,004 paired MCP--CLI trajectories, 1,889 ($94.3\%$) exhibit an
observed structured divergence. Figure~\ref{fig:additional_first_divergence}
further shows that the first divergence appears near the beginning of interaction
across model--environment strata, with semantic-action differences accounting for
most divergent pairs and realization differences forming a smaller but recurrent
component.

\ResultLead{Early divergence reshapes the remainder of the interaction trajectory.}
Figure~\ref{fig:supp_full_propagation} tracks the natural suffix after the first
consequential difference. Once the paired runs separate, they accumulate
different token costs and attempt counts and can reach different terminal
outcomes. The magnitude and direction of these changes vary across models and
environments. Figure~\ref{fig:additional_first_divergence}(c) adds a complementary
view. Semantic actions can align again later, but the extent of that later
agreement differs substantially across strata. Together, these results show how
a local interface difference changes the history conditioning subsequent
decisions, producing persistent, path-dependent differences in interaction cost
and task outcome.

\FloatBarrier
\ResultLead{Paired trajectories make the propagation mechanism concrete.}
Example~S1 shows a WebShop CLI quoting failure that forces a retry and redirects
subsequent interaction, while the paired MCP trajectory proceeds to task success.
Example~S2 shows the complementary pattern. An earlier ScienceWorld divergence
is followed by the same later \texttt{choose-0} semantic action, and both
trajectories ultimately succeed. Together, the examples illustrate two
complementary propagation patterns---divergence that persists into the endpoint
and divergence that is partially absorbed by later interaction.

\begingroup
\definecolor{DARTCaseBorder}{HTML}{DADDE0}
\definecolor{DARTCaseTitle}{HTML}{E3E4E5}
\definecolor{DARTCaseTask}{HTML}{EDF6EC}
\definecolor{DARTCaseContext}{HTML}{F5F5F7}
\definecolor{DARTCaseFailure}{HTML}{EEEEEE}
\definecolor{DARTCaseAction}{HTML}{FCF0F4}
\definecolor{DARTCaseOutcome}{HTML}{EDF4FC}
\newcommand{\DARTCase}[3]{\par\begingroup
  \fontencoding{T1}\fontsize{9.5}{11.7}\selectfont
  \setlength{\tabcolsep}{6pt}\renewcommand{\arraystretch}{1.00}%
  \setlength{\LTpre}{6pt}\setlength{\LTpost}{5pt}%
  \def\LTcaptype{appexample}%
  \DARTPrepareRows{#3}%
  \begin{longtable}{>{\bfseries\raggedright\arraybackslash}p{65pt}>{\raggedright\arraybackslash}p{\dimexpr\linewidth-65pt-24pt\relax}}
    \rowcolor{DARTCaseTitle}\multicolumn{2}{l}{\label{#1}\fontsize{10.2}{12.3}\selectfont\bfseries
      Example \theappexample.\enspace #2}\\[3pt]
    \endfirsthead
    \rowcolor{DARTCaseTitle}\multicolumn{2}{l}{\fontsize{10.2}{12.3}\selectfont\bfseries
      Example \theappexample.\enspace #2\ (continued)}\\[3pt]
    \endhead
    \endfoot
    \endlastfoot
    \DARTPreparedRows
  \end{longtable}
  \endgroup}
\DARTCase{ex:trajectory_realization}{WebShop --- a decode failure and retry}{
\rowcolor{DARTCaseTask}Task & Find me machine wash men\textquotesingle{}s dress shirts with cotton spandex, classic fit, short sleeve with color: delft, and size: x-large, and price lower than 70.00 dollars \\[2pt]
\rowcolor{DARTCaseContext}MCP search & \par \AppLiteral{\textless{}tool\_call\textgreater{}}\par \AppLiteral{\{\textquotedbl{}name\textquotedbl{}: \textquotedbl{}webshop\_\_search\textquotedbl{},}\par \hspace*{0.6em}\AppLiteral{\textquotedbl{}arguments\textquotedbl{}: \{\textquotedbl{}query\textquotedbl{}: \textquotedbl{}machine wash men\textquotesingle{}s dress shirt cotton spandex classic fit short sleeve delft x-large\textquotedbl{}\}\}}\par \AppLiteral{\textless{}/tool\_call\textgreater{}}\par \textbf{Observation:} \par [...]\par Page 1 (Total results: 50)\par [...]\par  \\[2pt]
\rowcolor{DARTCaseFailure}CLI failure & \par \AppLiteral{\textless{}command\textgreater{}}\par \AppLiteral{bench webshop search -{}-query \textquotesingle{}men\textquotesingle{}s dress shirts cotton spandex classic fit short sleeve delft x-large\textquotesingle{}}\par \AppLiteral{\textless{}/command\textgreater{}}\par  \\[2pt]
\rowcolor{DARTCaseFailure}CLI feedback & \par CLI validation error [surface.invalid\_argv] at surface\_decode. Constraint: surface\_response\_contract. invalid argv quoting: No closing quotation State transition: unchanged. Retryable: true. Allowed values: search, click. Correction: Current semantic operations: search, click.; Use exactly one current operation and all required arguments.; Return exactly one \textless{}command\textgreater{}argv command\textless{}/command\textgreater{} and no other text; each flag must have one value, and quote every multiword string value with POSIX single quotes.\par  \\[2pt]
\rowcolor{DARTCaseAction}CLI retry & \par \AppLiteral{\textless{}command\textgreater{}}\par \AppLiteral{bench webshop search -{}-query \textquotesingle{}mens dress shirts cotton spandex classic fit short sleeve delft x-large\textquotesingle{}}\par \AppLiteral{\textless{}/command\textgreater{}}\par \textbf{Observation:} \par [...]\par Page 1 (Total results: 50)\par [...]\par  \\[2pt]
\rowcolor{DARTCaseAction}MCP next action & \par \AppLiteral{\textless{}tool\_call\textgreater{}}\par \AppLiteral{\{\textquotedbl{}name\textquotedbl{}: \textquotedbl{}webshop\_\_click\textquotedbl{},}\par \hspace*{0.6em}\AppLiteral{\textquotedbl{}arguments\textquotedbl{}: \{\textquotedbl{}target\textquotedbl{}: \textquotedbl{}b085wqkrrj\textquotedbl{}\}\}}\par \AppLiteral{\textless{}/tool\_call\textgreater{}}\par \textbf{Observation:} \par [...]\par CLANMILUMS Men\textquotesingle{}s Classic Comfort Soft Regular Fit Short Sleeve Henley T-Shirt Tee\par Price: \$17.99 to \$19.99\par [...]\par  \\[2pt]
\rowcolor{DARTCaseOutcome}Outcome & \textbf{MCP:} task success.\par \textbf{CLI:} no task success; the environment-step limit is reached. \\[2pt]
}

\DARTCase{ex:trajectory_agreement}{ScienceWorld --- later action agreement}{
\rowcolor{DARTCaseTask}Task & Your task is to grow a orange plant from seed. Seeds can be found in the greenhouse. First, focus on a seed. Then, make changes to the environment that grow the plant until it reaches the reproduction life stage. \\[2pt]
\rowcolor{DARTCaseContext}Shared prefix & \textbf{MCP:}\par \AppLiteral{\textless{}tool\_call\textgreater{}}\par \AppLiteral{\{\textquotedbl{}name\textquotedbl{}: \textquotedbl{}scienceworld\_\_go\_to\textquotedbl{},}\par \hspace*{0.6em}\AppLiteral{\textquotedbl{}arguments\textquotedbl{}: \{\textquotedbl{}destination\textquotedbl{}: \textquotedbl{}greenhouse\textquotedbl{}\}\}}\par \AppLiteral{\textless{}/tool\_call\textgreater{}}\par \textbf{CLI:}\par \AppLiteral{\textless{}command\textgreater{}}\par \AppLiteral{bench scienceworld go-to -{}-destination \textquotesingle{}greenhouse\textquotesingle{}}\par \AppLiteral{\textless{}/command\textgreater{}}\par \textbf{Observation:} \par You move to the greenhouse.\par  \\[2pt]
\rowcolor{DARTCaseAction}MCP target & \par \AppLiteral{\textless{}tool\_call\textgreater{}}\par \AppLiteral{\{\textquotedbl{}name\textquotedbl{}: \textquotedbl{}scienceworld\_\_focus\_on\textquotedbl{},}\par \hspace*{0.6em}\AppLiteral{\textquotedbl{}arguments\textquotedbl{}: \{\textquotedbl{}target\textquotedbl{}: \textquotedbl{}orange seed in the seed stage\textquotedbl{}\}\}}\par \AppLiteral{\textless{}/tool\_call\textgreater{}}\par  \\[2pt]
\rowcolor{DARTCaseAction}CLI target & \par \AppLiteral{\textless{}command\textgreater{}}\par \AppLiteral{bench scienceworld focus-on -{}-target \textquotesingle{}orange seed in seed jar containing orange seed\textquotesingle{}}\par \AppLiteral{\textless{}/command\textgreater{}}\par  \\[2pt]
\rowcolor{DARTCaseAction}Environment & \par Ambiguous request: Please enter the number for the action you intended (or blank to cancel):\par 0:	focus on orange seed (in seed jar, in greenhouse)\par 1:	focus on orange seed (in seed jar, in greenhouse)\par 2:	focus on orange seed (in seed jar, in greenhouse)\par 3:	focus on orange seed (in seed jar, in greenhouse)\par 4:	focus on orange seed (in seed jar, in greenhouse)\par  \\[2pt]
\rowcolor{DARTCaseOutcome}Later action & \textbf{MCP:}\par \AppLiteral{\textless{}tool\_call\textgreater{}}\par \AppLiteral{\{\textquotedbl{}name\textquotedbl{}: \textquotedbl{}scienceworld\_\_choose\_0\textquotedbl{},}\par \hspace*{0.6em}\AppLiteral{\textquotedbl{}arguments\textquotedbl{}: \{\}\}}\par \AppLiteral{\textless{}/tool\_call\textgreater{}}\par \textbf{CLI:}\par \AppLiteral{\textless{}command\textgreater{}}\par \AppLiteral{bench scienceworld choose-0}\par \AppLiteral{\textless{}/command\textgreater{}}\par  \\[2pt]
\rowcolor{DARTCaseOutcome}Environment & \par You focus on the orange seed.\par  \\[2pt]
\rowcolor{DARTCaseOutcome}Outcome & Both trajectories eventually succeed. \\[2pt]
}
\endgroup

\subsection{RQ4: Commitment-Preserving Realization with DART}
\label{app:rq4_current_results}

\SVFigure{\includegraphics[width=\linewidth]{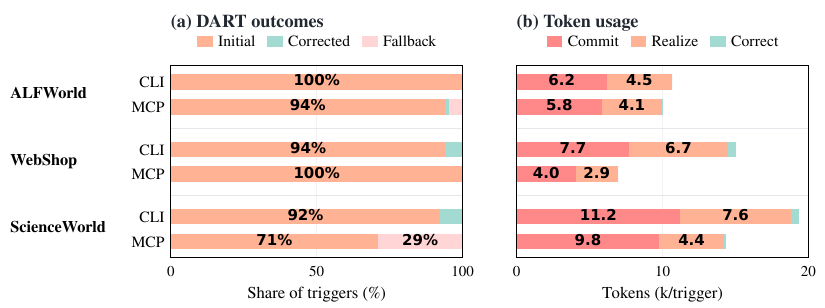}}{fig:supp_dart_accounting}{DART trigger outcomes
and token use across the six benchmark--interface settings evaluated in
Figure~\ref{fig:dart_policy}. (a) Share of triggers admitted on the initial
realization, after preserving correction, or released to standard retry. (b) Mean
tokens per trigger spent on semantic commitment, realization, and correction.}

\ResultLead{DART concentrates recovery at the realization boundary while preserving the validated semantic commitment.}
Across the six benchmark--interface settings,
71--100\% of triggered interventions are admitted on the first realization.
Among the four interventions that proceed to a preserving-correction
opportunity, three are admitted after that single correction. These results
show that DART resolves triggered failures primarily by repairing the
interface-specific expression while keeping the committed operation and typed
arguments fixed.

\FloatBarrier
\ResultLead{Recorded trajectories instantiate both successful DART paths.}
Example~S3 shows direct realization. After an ALFWorld/MCP failure, DART commits
to \texttt{alfworld.move}, re-expresses the same typed action in a valid tool
call, and executes it successfully. Example~S4 shows preserving correction. The
first WebShop/CLI realization loses the literal apostrophes in
\texttt{18''x18''}, while the correction changes only the command expression,
preserves \texttt{webshop.click} and the full target value, and is admitted before
the agent completes the purchase. These trajectories provide concrete executions
of the two success branches summarized in Figure~\ref{fig:supp_dart_accounting}.

\begingroup
\definecolor{DARTCaseBorder}{HTML}{DADDE0}
\definecolor{DARTCaseTitle}{HTML}{E3E4E5}
\definecolor{DARTCaseTask}{HTML}{EDF6EC}
\definecolor{DARTCaseContext}{HTML}{F5F5F7}
\definecolor{DARTCaseFailure}{HTML}{EEEEEE}
\definecolor{DARTCaseAction}{HTML}{FCF0F4}
\definecolor{DARTCaseOutcome}{HTML}{EDF4FC}
\newcommand{\DARTCase}[3]{\par\begingroup
  \fontencoding{T1}\fontsize{9.5}{11.7}\selectfont
  \setlength{\tabcolsep}{6pt}\renewcommand{\arraystretch}{1.00}%
  \setlength{\LTpre}{6pt}\setlength{\LTpost}{5pt}%
  \def\LTcaptype{appexample}%
  \DARTPrepareRows{#3}%
  \begin{longtable}{>{\bfseries\raggedright\arraybackslash}p{65pt}>{\raggedright\arraybackslash}p{\dimexpr\linewidth-65pt-24pt\relax}}
    \rowcolor{DARTCaseTitle}\multicolumn{2}{l}{\label{#1}\fontsize{10.2}{12.3}\selectfont\bfseries
      Example \theappexample.\enspace #2}\\[3pt]
    \endfirsthead
    \rowcolor{DARTCaseTitle}\multicolumn{2}{l}{\fontsize{10.2}{12.3}\selectfont\bfseries
      Example \theappexample.\enspace #2\ (continued)}\\[3pt]
    \endhead
    \endfoot
    \endlastfoot
    \DARTPreparedRows
  \end{longtable}
  \endgroup}
\DARTCase{ex:dart_mcp_admission}{ALFWorld / MCP --- realizing a validated action}{
\rowcolor{DARTCaseTask}Task & cool some mug and put it in coffeemachine. \\[2pt]
\rowcolor{DARTCaseContext}Recent context & \textbf{Executed:} \AppLiteral{cool mug 1 with fridge 1}\par \textbf{Observation:} You cool the mug 1 using the fridge 1.\par \textbf{Executed:} \AppLiteral{go to coffeemachine 1}\par \textbf{Observation:} You arrive at coffeemachine 1. On the coffeemachine 1, you see nothing. \\[2pt]
\rowcolor{DARTCaseFailure}Failed output & \par \AppLiteral{\textless{}tool\_call\textgreater{}}\par \AppLiteral{\{\textquotedbl{}name\textquotedbl{}: \textquotedbl{}alfworld\_\_move\textquotedbl{},}\par \hspace*{0.6em}\AppLiteral{\textquotedbl{}destination\textquotedbl{}: \textquotedbl{}coffeemachine 1\textquotedbl{},}\par \hspace*{0.6em}\AppLiteral{\textquotedbl{}object\textquotedbl{}: \textquotedbl{}mug 1\textquotedbl{}\}}\par \AppLiteral{\textless{}/tool\_call\textgreater{}}\par  \\[2pt]
\rowcolor{DARTCaseFailure}Feedback & \par tool call must contain exactly name and arguments\par The proposal is rejected before environment execution. \\[2pt]
\rowcolor{DARTCaseAction}Commitment & \par \AppLiteral{\{\textquotedbl{}schema\_id\textquotedbl{}: \textquotedbl{}alfworld.move\textquotedbl{},}\par \hspace*{0.6em}\AppLiteral{\textquotedbl{}arguments\textquotedbl{}: \{\textquotedbl{}object\textquotedbl{}: \textquotedbl{}mug 1\textquotedbl{},}\par \hspace*{9em}\AppLiteral{\textquotedbl{}destination\textquotedbl{}: \textquotedbl{}coffeemachine 1\textquotedbl{}\}\}}\par  \\[2pt]
\rowcolor{DARTCaseAction}Realization & \par \AppLiteral{\textless{}tool\_call\textgreater{}}\par \AppLiteral{\{\textquotedbl{}name\textquotedbl{}: \textquotedbl{}alfworld\_\_move\textquotedbl{},}\par \hspace*{0.6em}\AppLiteral{\textquotedbl{}arguments\textquotedbl{}: \{\textquotedbl{}destination\textquotedbl{}: \textquotedbl{}coffeemachine 1\textquotedbl{},}\par \hspace*{9em}\AppLiteral{\textquotedbl{}object\textquotedbl{}: \textquotedbl{}mug 1\textquotedbl{}\}\}}\par \AppLiteral{\textless{}/tool\_call\textgreater{}}\par Admitted on the first DART realization; no correction call is needed. \\[2pt]
\rowcolor{DARTCaseOutcome}Execution & \textbf{Executed:} \AppLiteral{move mug 1 to coffeemachine 1}\par \textbf{Observation:} You move the mug 1 to the coffeemachine 1.\par \textbf{Recorded outcome:} task success, score 1.0. \\[2pt]
}

\DARTCase{ex:dart_cli_correction}{WebShop / CLI --- preserving a quoted option value}{
\rowcolor{DARTCaseTask}Task & Find me super soft decorative pillows for living room with color: lavender, and size: 18\textquotesingle{}\textquotesingle{}x18\textquotesingle{}\textquotesingle{}, and price lower than 50.00 dollars \\[2pt]
\rowcolor{DARTCaseContext}Current page & \par You have clicked lavender.\par [...]\par size\par   [button] 18\textquotesingle{}\textquotesingle{}x18\textquotesingle{}\textquotesingle{} [button\_]\par   [button] 20\textquotesingle{}\textquotesingle{}x20\textquotesingle{}\textquotesingle{} [button\_]\par [...]\par Price: \$13.99\par [...]\par [button] Buy Now [button\_]\par  \\[2pt]
\rowcolor{DARTCaseFailure}Failed output & \par \AppLiteral{\textless{}command\textgreater{}}\par \AppLiteral{bench webshop click -{}-target \textquotesingle{}18\textquotesingle{}\textquotesingle{}x18\textquotesingle{}\textquotesingle{}\textquotesingle{}}\par \AppLiteral{\textless{}/command\textgreater{}}\par  \\[2pt]
\rowcolor{DARTCaseFailure}Feedback & \par target is not uniquely available in the current state\par The parsed target is \AppLiteral{18x18}; its literal apostrophes have been lost. \\[2pt]
\rowcolor{DARTCaseAction}Commitment & \par \AppLiteral{\{\textquotedbl{}schema\_id\textquotedbl{}: \textquotedbl{}webshop.click\textquotedbl{},}\par \hspace*{0.6em}\AppLiteral{\textquotedbl{}arguments\textquotedbl{}: \{\textquotedbl{}target\textquotedbl{}: \textquotedbl{}18\textquotesingle{}\textquotesingle{}x18\textquotesingle{}\textquotesingle{}\textquotedbl{}\}\}}\par  \\[2pt]
\rowcolor{DARTCaseAction}First realization & \par \AppLiteral{\textless{}command\textgreater{}}\par \AppLiteral{bench webshop click -{}-target \textquotesingle{}18\textquotesingle{}\textquotesingle{}x18\textquotesingle{}\textquotesingle{}\textquotesingle{}}\par \AppLiteral{\textless{}/command\textgreater{}}\par \textbf{Feedback:} target is not uniquely available in the current state\par The target is still \AppLiteral{18x18}, so this realization is not admitted. \\[2pt]
\rowcolor{DARTCaseAction}Correction & \par \AppLiteral{\textless{}command\textgreater{}}\par \AppLiteral{bench webshop click}\par \AppLiteral{-{}-target \textquotesingle{}18\textquotesingle{}\textquotedbl{}\textquotesingle{}\textquotedbl{}\textquotesingle{}\textquotesingle{}\textquotedbl{}\textquotesingle{}\textquotedbl{}\textquotesingle{}x18\textquotesingle{}\textquotedbl{}\textquotesingle{}\textquotedbl{}\textquotesingle{}\textquotesingle{}\textquotedbl{}\textquotesingle{}\textquotedbl{}\textquotesingle{}\textquotesingle{}}\par \AppLiteral{\textless{}/command\textgreater{}}\par Admitted: the parsed target now matches the committed \AppLiteral{18\textquotesingle{}\textquotesingle{}x18\textquotesingle{}\textquotesingle{}}. \\[2pt]
\rowcolor{DARTCaseOutcome}Environment & \par You have clicked 18\textquotesingle{}\textquotesingle{}x18\textquotesingle{}\textquotesingle{}.\par You have clicked lavender.\par [...]\par  \\[2pt]
\rowcolor{DARTCaseOutcome}Next action & \par \AppLiteral{\textless{}command\textgreater{}}\par \AppLiteral{bench webshop click -{}-target \textquotesingle{}buy now\textquotesingle{}}\par \AppLiteral{\textless{}/command\textgreater{}}\par \textbf{Observation:} \par Thank you for shopping with us!\par [...]\par Your score (min 0.0, max 1.0)\par 1.0\par [...]\par  \\[2pt]
}
\endgroup

\ResultLead{DART localizes its additional computation to commitment and realization.}
The token decomposition in Figure~\ref{fig:supp_dart_accounting}(b) shows that
commitment formation is the larger component of DART token use in every
evaluated stratum, followed by realization, while correction is invoked only for
the small subset of triggers that require a second surface attempt. This targeted
accounting complements the episode-level results in Figure~\ref{fig:dart_policy}.
Under MCP, DART uses fewer total tokens in all three environments. DART improves
benchmark score across every evaluated benchmark--interface setting.

\endgroup
\FloatBarrier
\renewcommand{\FloatBarrier}{\par}

\newcommand{\ProtocolLead}[1]{\par\noindent\textbf{#1}\ }

\section{Experimental Protocol and Statistical Analysis}
\label{app:evaluation_protocol}

This appendix specifies the controlled evaluation protocol underlying MICA.

\subsection{Evaluation Suite and Controlled Comparison}
\label{app:protocol}
\label{app:benchmark_metrics}

\ProtocolLead{Evaluation suite.}
The baseline study evaluates six models---Qwen3-4B, Qwen3-8B, Qwen3-14B,
Qwen3-32B \citep{yang2025qwen3}, DeepSeek-v4-Flash, and
Gemini-2.5-Flash---on three interactive environments. Each model is evaluated
on 134 ALFWorld tasks, 100 WebShop tasks, and 100 ScienceWorld tasks, yielding
2,004 paired MCP--CLI episodes. ALFWorld is evaluated by task success. WebShop
uses its native reward and ScienceWorld its native benchmark score, with the
latter two reported on a 0--100 scale in the paper. Interaction cost is the
total recorded input-plus-output model tokens consumed along a trajectory,
including retry and intervention requests. Native benchmark interfaces are
retained as calibration references for endpoint behavior.

\ProtocolLead{Controlled interface comparison.}
For every paired task, MCP and CLI use the same model checkpoint, initial task
state, public task information, semantic action contract, shared validation and
execution path, generation configuration, and interaction budget. The controlled
variation lies in capability presentation, action realization, and feedback
representation.

\Needspace{4\baselineskip}
\subsection{Generation, Context, and Interaction Budgets}
\phantomsection\label{app:current_configuration}

\ProtocolLead{Generation settings.}
Natural episode evaluation uses temperature-zero decoding, whereas matched-state
probes use $T=0.6$ to estimate local response distributions from repeated
samples. All requests use top-$p=1$, a 24,576-token input ceiling, and a
16,384-token output ceiling. The four Qwen3 models are evaluated from local
checkpoints. DeepSeek-v4-Flash and Gemini-2.5-Flash use their corresponding API
bindings under the same registered protocol.

\ProtocolLead{Context and episode budgets.}
Interaction histories are retained as complete events and truncated oldest-first
to fixed benchmark-specific limits of 100 events / 24,000 tokens for ALFWorld,
30 / 12,000 for WebShop, and 8 / 8,000 for ScienceWorld. The immediately
preceding failed expression and its public feedback remain in context for a
correction request. Episode-level environment-action / decision / model-call
ceilings are 50 / 75 / 150 for ALFWorld, 30 / 45 / 90 for WebShop, and
32 / 32 / 32 for ScienceWorld. The same ceilings apply to MCP and CLI within
each paired comparison.

Every dispatched model request contributes to interaction cost. A proposal
rejected before execution counts as an action attempt but does not consume an
environment transition. Under the Base policy, a decision receives at most one
correction opportunity. When DART is activated, semantic selection, realization,
and at most one preserving correction all operate within the remaining
episode-level call budget.
\phantomsection\label{app:context_policy}
\FloatBarrier

\subsection{Paired Episode Evaluation Protocol}
\label{app:trajectory_analysis}
\label{app:randomness}

\ProtocolLead{Matched episode construction.}
MCP and CLI episodes are paired by task identifier and initial state under the
same checkpoint, public task information, generation configuration, registered
randomness policy, and episode budget. Complete replicates are averaged within
task before paired aggregation. An episode terminates on native task completion,
an explicit terminal response, or the corresponding registered interaction
budget.

\ProtocolLead{Common prefix and first divergence.}
For each pair, we align the longest semantic prefix over which the normalized
proposal, validation result, execution outcome, public feedback, and recovery
record agree. Surface strings may differ within this prefix when they decode to
the same semantic action and produce the same interaction outcome. The first
step at which this interaction signature differs defines the first consequential
divergence $t^\star$ used throughout the trajectory analysis.

\ProtocolLead{Natural propagation.}
After $t^\star$, each episode continues under its own interaction history and
environment state. We characterize the resulting suffix through subsequent
semantic-action agreement, failure and recovery events, charged tokens,
interaction attempts, and terminal outcome. Paired natural trajectories thus
connect a common initial condition to the downstream effects summarized in
Appendix~\ref{app:rq3_current_results}.

\subsection{Matched-State Probe Protocol}
\label{app:matched_probes}
\label{app:matched_sampling}

\ProtocolLead{Probe design.}
The matched-state experiment contains \NProbeStates{} source states and
\NProbeResponses{} sampled responses across all six models and three
environments. Each source state is evaluated under the eight
$(\Phi,R,\Psi)$ assignments in Table~\ref{tab:factorial}, with eight independent
draws per condition. Every draw restarts from the same registered source
context, so repeated sampling estimates the local response distribution without
allowing one draw to alter the state observed by the next.

\ProtocolLead{Matched decision contexts.}
Source states and public histories are fixed before sampling using the
registered category/hash selection rule. Within a source state, the environment
state, policy-visible history, semantic contract, model, and generation
configuration are held fixed while capability presentation $\Phi$, realization
protocol $R$, and feedback representation $\Psi$ are varied. For ScienceWorld,
the matched context is the recorded policy-visible source state and public
interaction history supplied to the model.

\phantomsection\label{app:factorized_context}
\ProtocolLead{Factorial interface assignment.}
Presentation specifies operation descriptions, argument types, and public
constraints. Realization specifies the response container, operation binding,
argument serialization, and decoder. Feedback specifies how normalized public
results and failure facts are rendered. Mixed assignments use a canonical
semantic ledger to render prior decoded actions and public results under the
assigned components, preserving the semantic interaction history while changing
the representation exposed to the model.

\begin{AppTable}
\caption{\textbf{Factorial interface assignments used in the matched-state probes.}}
\label{tab:factorial}
\centering
\small
\renewcommand{\arraystretch}{1.03}
\begin{tabular}{@{}lccc@{}}
\toprule
\textbf{Condition} & \textbf{Presentation $\Phi$} & \textbf{Realization $R$} & \textbf{Feedback $\Psi$} \\
\midrule
C1 & MCP & MCP & MCP \\
C2 & MCP & MCP & CLI \\
C3 & MCP & CLI & MCP \\
C4 & MCP & CLI & CLI \\
C5 & CLI & MCP & MCP \\
C6 & CLI & MCP & CLI \\
C7 & CLI & CLI & MCP \\
C8 & CLI & CLI & CLI \\
\bottomrule
\end{tabular}
\end{AppTable}

\phantomsection\label{app:rq3_eight_draw_design}
C1 and C8 recover the homogeneous MCP and CLI interfaces, while C2--C7 expose
component substitutions at the same decision context. These eight conditions
form the response profiles underlying the component and interaction effects in
Figure~\ref{fig:factorial_effects}.
\FloatBarrier

\Needspace{11\baselineskip}
\subsection{Metrics and Statistical Analysis}
\label{app:metrics}
\label{app:statistical-inference}

\subsubsection{Paired Episode Estimands}
\label{app:paired_estimands}

\ProtocolLead{Paired effect.}
Tasks are the unit of paired episode comparison. For metric $m$, $N$ tasks in a
model--environment stratum, and $K_i$ registered replicate pairs for task $i$,
the MCP-minus-CLI effect is
\begin{equation}
    \Delta_m
    =
    \frac{1}{N}\sum_{i=1}^{N}\frac{1}{K_i}
    \sum_{r=1}^{K_i}
    \left(
    m_{ir}^{\mathrm{MCP}}-m_{ir}^{\mathrm{CLI}}
    \right).
    \label{eq:paired_effect}
\end{equation}
Replicates are averaged within task before aggregation, so each task receives
equal weight within a stratum. We report pointwise percentile 95\% intervals
from 10,000 paired task-cluster bootstrap resamples
\citep{efron1979bootstrap}. RQ4 uses the same task-clustered procedure within
each protocol-matched Base--DART comparison. Positive outcome contrasts denote
larger MCP outcomes. Positive token, call, attempt, or step contrasts denote
larger MCP resource use.

\subsubsection{Failure and Recovery Metrics}
\label{app:failure_recovery_metrics}
\phantomsection\label{app:current_rq2_metrics}

\ProtocolLead{Failure metrics.}
Failure incidence is measured over all action attempts, including correction
attempts. Failure composition is measured over failed attempts and assigns each
attempt to its earliest registered failure boundary in the action pipeline.
Semantic-choice profiles are computed from decoded normalized proposals,
separating changes in action selection from failures of surface realization.

\ProtocolLead{Recovery metrics.}
Recovery analysis begins from an eligible first-failure root, a retryable,
policy-owned failure whose environment state remains unchanged. The immediate
successor is classified by whether the model changes its semantic proposal,
produces an admitted repair, repeats the same error, produces a different
failure, or does not retry. Retry acceptance measures whether that successor is
admitted, while later task success records the eventual episode outcome. Retry
cost is the charged token use of the successor response. Bootstrap resampling
keeps all failure roots from the same task together, preserving the task as the
statistical resampling unit.

\subsubsection{Matched-State Response Statistics}
\label{app:matched_estimators}

\ProtocolLead{Semantic response distribution.}
At a matched decision context, each draw yields either a realization failure or
a decoded semantic proposal with a shared-validation outcome. Let $n_t^I(a)$
denote the number of decoded draws proposing semantic action $a$,
$n_{t,\mathrm{dec}}^I=\sum_a n_t^I(a)$ the number of decoded draws, and
$n_t^I$ the total number of draws. The empirical semantic-action distribution
and its entropy are
\begin{equation}
 \widehat p_t^I(a\mid\mathrm{dec})
 =
 \frac{n_t^I(a)}{n_{t,\mathrm{dec}}^I},
 \qquad
 H_t^I
 =
 -\sum_a
 \widehat p_t^I(a\mid\mathrm{dec})
 \log_2 \widehat p_t^I(a\mid\mathrm{dec}).
 \label{eq:semantic_uncertainty}
\end{equation}
Realization failure is the complement of decoding coverage, while admission is
the fraction of decoded proposals accepted by the shared semantic validator.
Action entropy summarizes the dispersion of decoded semantic choices and is
reported for conditions with at least two decoded draws.

\ProtocolLead{Probe aggregation.}
Realization-failure and entropy summaries weight source tasks equally.
Admission pools admitted and decoded draws within each condition. Bootstrap
resamples source tasks jointly across C1--C8, preserving all factorial
assignments from a source state in the same resample. Component main effects
compare the MCP and CLI levels of one interface block while averaging over the
assignments of the remaining blocks, matching the factorial contrasts reported
in the main text.

\newcommand{\ConstructionLead}[1]{\par\noindent\textbf{#1}\ }

\section{Contract-Matched Interface Construction}
\label{app:supplement}
\label{app:formal}
\label{app:interfaces}
\label{app:implementation}

This appendix details how MCP and CLI are constructed from a common semantic
action contract. The construction separates the shared executable semantics
from the model-facing representation. Both interfaces expose the same
state-local operations and typed arguments, decode back to the same normalized
action space, and enter the same validation, compilation, and execution path.

\subsection{Shared Semantic Boundary}

\ConstructionLead{Semantic actions define the common unit of comparison.}
A semantic action $a=(u,v)$ consists of an operation $u$ and its complete tuple
of normalized typed arguments $v$. At environment state $s_t$, the public
contract $\mathcal C_t$ specifies the operations currently exposed to the
policy, their argument domains, and their native execution bindings. MCP and
CLI render this same contract through different surface forms, but both decode
model responses back to the contract's semantic action space before shared
validation. Two surface responses are therefore treated as semantically
equivalent at this boundary whenever they specify the same registered operation
and equivalent normalized arguments.

\ConstructionLead{Interface components alter representation, not execution semantics.}
Following the factorization in the main text, presentation $\Phi_I$ determines
how available capabilities and constraints are shown to the model, realization
$R_I=(\rho_I,D_I)$ determines the response grammar and decoder, and feedback
$\Psi_I$ determines how public execution results and failure information are
rendered. These components can change what the model sees and how it expresses
an action, while the semantic validator, native compiler, and environment
executor remain shared across interfaces.

\subsection{Surface Rendering and Conformance Checks}
\label{app:interface_conformance}

\ConstructionLead{Shared contract, distinct surfaces.}
For every semantic operation, the MCP renderer produces a named tool with a
structured argument object, whereas the CLI renderer produces a registered
command with explicit flags and argument values. The CLI decoder parses the
registered argument vector directly. Quoted multiword values remain a single
typed argument. Both renderers are generated from $\mathcal C_t$, so differences
in naming, serialization, quoting, and response containers change the surface
representation without introducing new executable capabilities.

\ConstructionLead{Round-trip semantic equivalence.}
Conformance first checks that a registered semantic action survives rendering
and decoding under either interface,
\begin{equation}
D_{\mathrm{MCP}}\!\left(
    \operatorname{Render}_{\mathrm{MCP}}(a)
\right)
\equiv_{\mathrm{sem}}
a
\equiv_{\mathrm{sem}}
D_{\mathrm{CLI}}\!\left(
    \operatorname{Render}_{\mathrm{CLI}}(a)
\right).
\label{eq:roundtrip_conformance_app}
\end{equation}
Semantic equivalence requires agreement on the operation and every normalized
typed argument. Surface-level differences such as whitespace, JSON field order,
or valid CLI quoting therefore leave the decoded action unchanged.

\ConstructionLead{Common native compilation.}
A decoded proposal that passes the shared semantic validator is compiled by the
same benchmark-specific compiler $K_{\mathcal C_t}$. For semantically matched
MCP and CLI responses,
\begin{equation}
 K_{\mathcal C_t}\!\left(D_{\mathrm{MCP}}(y_{\mathrm{MCP}})\right)
 =
 K_{\mathcal C_t}(a)
 =
 K_{\mathcal C_t}\!\left(D_{\mathrm{CLI}}(y_{\mathrm{CLI}})\right).
 \label{eq:shared_compilation_app}
\end{equation}
Conformance checks combine four tests, namely coverage of the registered semantic
actions, round-trip decoding, common native compilation, and consistency of the
policy-visible contract facts. Finite state-local action catalogs are checked
over their registered members. Open string arguments are checked over the
declared benchmark domain and schema-valid test cases.

\subsection{Benchmark-Specific Contract Realizations}
\label{app:action_mappings}
\label{app:environments}
\label{app:semantic_mapping_examples}

\ConstructionLead{ALFWorld.}
For ALFWorld~\citep{shridhar2021alfworld}, the state-local contract is built
from the currently admissible native action instances. Operations such as
navigation, pickup, placement, opening, closing, heating, cooling, and cleaning
retain their complete object and destination names as typed arguments. MCP and
CLI therefore expose different realizations of the same currently admissible
action instances.

\ConstructionLead{WebShop.}
WebShop~\citep{yao2022webshop} is represented by two semantic operations, namely
\texttt{search(query)} and \texttt{click(target)}. The search argument is an
open string satisfying the shared schema, whereas click targets are drawn from
the controls currently visible to the agent. Product options, pagination
controls, and purchase actions are represented as values of
\texttt{click(target)} rather than as additional interface-specific operations.

\ConstructionLead{ScienceWorld.}
ScienceWorld~\citep{wang2022scienceworld} lifts the benchmark's official action
templates into typed semantic operations and argument tuples. Public object
referents, aliases, and tuple structure determine the state-local arguments
available to the policy. MCP and CLI render these typed operations differently
but compile admitted proposals back to the same native ScienceWorld action.

The following compact mappings illustrate the construction used across the
three environments. In each block, the MCP and CLI surfaces decode to the
displayed semantic action and compile to the same native command.

\begingroup
\small
\setlength{\LTpre}{4pt}\setlength{\LTpost}{4pt}%
\def\LTcaptype{appmapping}%
\begin{longtable}{@{}L{.1748\linewidth}L{\dimexpr.7452\linewidth-2\tabcolsep\relax}@{}}
\multicolumn{2}{@{}l}{\textbf{ALFWorld}}\\[2pt]
\toprule
\textbf{Layer} & \textbf{Representation} \\
\midrule
\endfirsthead
\multicolumn{2}{@{}l}{\textbf{ALFWorld (continued)}}\\[2pt]
\toprule
\textbf{Layer} & \textbf{Representation} \\
\midrule
\endhead
\endfoot
\endlastfoot
Semantic action &
\texttt{\detokenize{alfworld.go(destination="desk 1")}} \\
MCP surface &
\texttt{\detokenize{<tool_call>}}\par
\texttt{\{"name":"alfworld\_\_go",}\par
\hspace*{1em}\texttt{"arguments":\{"destination":"desk 1"\}\}}\par
\texttt{\detokenize{</tool_call>}} \\
CLI surface &
\texttt{\detokenize{<command>}}\par
\texttt{\detokenize{bench alfworld go --destination 'desk 1'}}\par
\texttt{\detokenize{</command>}} \\
Native action &
\texttt{\detokenize{go to desk 1}} \\
\bottomrule
\end{longtable}

\begin{longtable}{@{}L{.1748\linewidth}L{\dimexpr.7452\linewidth-2\tabcolsep\relax}@{}}
\multicolumn{2}{@{}l}{\textbf{WebShop}}\\[2pt]
\toprule
\textbf{Layer} & \textbf{Representation} \\
\midrule
\endfirsthead
\multicolumn{2}{@{}l}{\textbf{WebShop (continued)}}\\[2pt]
\toprule
\textbf{Layer} & \textbf{Representation} \\
\midrule
\endhead
\endfoot
\endlastfoot
Semantic action &
\texttt{\detokenize{webshop.click(target="next >")}} \\
MCP surface &
\texttt{\detokenize{<tool_call>}}\par
\texttt{\{"name":"webshop\_\_click",}\par
\hspace*{1em}\texttt{"arguments":\{"target":"next >"\}\}}\par
\texttt{\detokenize{</tool_call>}} \\
CLI surface &
\texttt{\detokenize{<command>}}\par
\texttt{\detokenize{bench webshop click --target 'next >'}}\par
\texttt{\detokenize{</command>}} \\
Native action &
\texttt{\detokenize{click[next >]}} \\
\bottomrule
\end{longtable}

\begin{longtable}{@{}L{.1748\linewidth}L{\dimexpr.7452\linewidth-2\tabcolsep\relax}@{}}
\multicolumn{2}{@{}l}{\textbf{ScienceWorld}}\\[2pt]
\toprule
\textbf{Layer} & \textbf{Representation} \\
\midrule
\endfirsthead
\multicolumn{2}{@{}l}{\textbf{ScienceWorld (continued)}}\\[2pt]
\toprule
\textbf{Layer} & \textbf{Representation} \\
\midrule
\endhead
\endfoot
\endlastfoot
Semantic action &
\texttt{\detokenize{scienceworld.t2.go_to(destination="hallway")}} \\
MCP surface &
\texttt{\detokenize{<tool_call>}}\par
\texttt{\{"name":"scienceworld\_\_go\_to",}\par
\hspace*{1em}\texttt{"arguments":\{"destination":"hallway"\}\}}\par
\texttt{\detokenize{</tool_call>}} \\
CLI surface &
\texttt{\detokenize{<command>}}\par
\texttt{\detokenize{bench scienceworld go-to --destination hallway}}\par
\texttt{\detokenize{</command>}} \\
Native action &
\texttt{\detokenize{go to hallway}} \\
\bottomrule
\end{longtable}
\endgroup

\subsection{Worked Round-Trip Example}
\label{app:illustrative_case}

\ConstructionLead{A single semantic action admits two surface realizations.}
Consider an ALFWorld state whose contract includes
\texttt{\detokenize{alfworld.go(destination="sofa 1")}}. The two interfaces
present and realize this action differently, but both responses round-trip to
the same typed proposal and therefore enter the same shared validation and
execution path.

\begingroup
\definecolor{DARTCaseBorder}{HTML}{DADDE0}
\definecolor{DARTCaseTitle}{HTML}{E3E4E5}
\definecolor{DARTCaseTask}{HTML}{EDF6EC}
\definecolor{DARTCaseContext}{HTML}{F5F5F7}
\definecolor{DARTCaseFailure}{HTML}{EEEEEE}
\definecolor{DARTCaseAction}{HTML}{FCF0F4}
\definecolor{DARTCaseOutcome}{HTML}{EDF4FC}
\newcommand{\DARTCase}[3]{\par\begingroup
  \fontencoding{T1}\fontsize{9.5}{11.7}\selectfont
  \setlength{\tabcolsep}{6pt}\renewcommand{\arraystretch}{1.00}%
  \setlength{\LTpre}{6pt}\setlength{\LTpost}{5pt}%
  \def\LTcaptype{appexample}%
  \DARTPrepareRows{#3}%
  \begin{longtable}{>{\bfseries\raggedright\arraybackslash}p{65pt}>{\raggedright\arraybackslash}p{\dimexpr\linewidth-65pt-24pt\relax}}
    \rowcolor{DARTCaseTitle}\multicolumn{2}{l}{\label{#1}\fontsize{10.2}{12.3}\selectfont\bfseries
      Example \theappexample.\enspace #2}\\[3pt]
    \endfirsthead
    \rowcolor{DARTCaseTitle}\multicolumn{2}{l}{\fontsize{10.2}{12.3}\selectfont\bfseries
      Example \theappexample.\enspace #2\ (continued)}\\[3pt]
    \endhead
    \endfoot
    \endlastfoot
    \DARTPreparedRows
  \end{longtable}
  \endgroup}
\DARTCase{ex:repair_case}{Round-trip equivalence for an ALFWorld action}{
\rowcolor{DARTCaseTask}Semantic action &
\AppLiteral{alfworld.go(destination="sofa 1")} \\[2pt]

\rowcolor{DARTCaseAction}MCP realization &
\AppLiteral{\textless{}tool\_call\textgreater{}}
\par\AppLiteral{\{"name":"alfworld\_\_go",}
\par\hspace*{.6em}\AppLiteral{"arguments":\{"destination":"sofa 1"\}\}}
\par\AppLiteral{\textless{}/tool\_call\textgreater{}} \\[2pt]

\rowcolor{DARTCaseAction}CLI realization &
\AppLiteral{\textless{}command\textgreater{}}
\par\AppLiteral{bench alfworld go -{}-destination \textquotesingle{}sofa 1\textquotesingle{}}
\par\AppLiteral{\textless{}/command\textgreater{}} \\[2pt]

\rowcolor{DARTCaseOutcome}Decoded action &
Both decoders return
\AppLiteral{alfworld.go(destination="sofa 1")}. \\[2pt]

\rowcolor{DARTCaseOutcome}Compilation &
Both proposals compile to
\AppLiteral{go to sofa 1}. \\[2pt]
}
\endgroup

The two surface forms decode to the same semantic action and compile to the
same native command. Appendix~\ref{app:matched_probes}
uses this shared boundary to exchange interface components at matched decision
contexts, while Appendix~\ref{app:dart_final_protocol} uses the same semantic
identity criterion to preserve validated actions during targeted realization
repair.

\newcommand{\DARTLead}[1]{\par\noindent\textbf{#1}\ }

\section{DART: Controller, Triggers, and Evaluation Protocol}
\label{app:dart_implementation}

This appendix specifies the controller used in RQ4. DART separates semantic
selection from interface-specific realization, binds a validated semantic action
as an explicit commitment, and restricts subsequent repair to surface
realization while that commitment remains valid.

\subsection{Commitment-Preserving Realization}
\label{app:dart}
\label{app:dart_controls}

\DARTLead{Commitment formation.}
After an eligible failure, DART asks the same underlying model to select a
canonical semantic action under the current public contract. A decoded selection
that passes the shared semantic validator establishes
$c_t=(\kappa_t,\bar a_t)$, where $\kappa_t$ identifies the contract and
$\bar a_t=(\bar u_t,\bar v_t)$ contains the normalized operation and all typed
arguments. This explicit commitment becomes the semantic target for the
interface-specific realization stage.

\DARTLead{Commitment-preserving realization.}
A realization is admitted only when the active contract is unchanged, the
surface response decodes to the committed semantic action, and that action
remains valid in the current state,
\begin{equation}
 \kappa_{\mathrm{current}}=\kappa_t,\qquad
 D_I(y_t^I)\equiv_{\kappa_t}\bar a_t,\qquad
 V(s_t,\bar a_t)=1.
 \label{eq:dart_full_gate}
\end{equation}
The semantic-equivalence test requires the same operation and every normalized
typed argument. Thus, DART permits the interface expression to change while
holding the selected action fixed.

\DARTLead{Repair boundary.}
If the initial realization fails and the commitment remains valid, DART issues
one preserving-correction request containing the validated commitment, the
failed surface expression, and its public failure feedback. The correction may
change quoting, serialization, container structure, or other
interface-specific expression details, but not the committed operation or
arguments. If the contract changes, the commitment becomes invalid, or the
repair budget is exhausted, control returns to the ordinary retry policy.

\subsection{Error-Triggered Controller}

\DARTLead{Controller flow.}
Algorithm~\ref{alg:dart_current} gives the commitment-preserving branch used
after a registered trigger. Semantic selection and initial realization require
two model calls. A preserving correction is attempted only when a third
episode-level call slot remains available. The
$\operatorname{Invoke}$ routine enforces the active episode-level call budget
and per-request limits before every dispatch.

\begin{AppAlgorithm}{alg:dart_current}{Error-triggered commitment-preserving repair}
\AppIO{Public retry context $x$; failure $f$; contract $\mathcal C$; interface $I$; remaining call budget $B_{\mathrm{call}}$.}
{An admitted proposal, or $\operatorname{Release}$ to the ordinary retry policy.}

\AppLine{0}{\textbf{if} $\neg\operatorname{Eligible}(f)$ or $B_{\mathrm{call}}<2$ \textbf{then return} $\operatorname{Release}$}

\AppLine{0}{$z\gets\operatorname{Invoke}(\mathrm{selection},x,f,\mathcal C)$}

\AppLine{0}{\textbf{if} $z=\bot$ \textbf{then return} $\operatorname{Release}$}

\AppLine{0}{$\bar a\gets\operatorname{DecodeSelection}(z,\mathcal C)$}

\AppLine{0}{\textbf{if} $\bar a=\bot$ or $\neg V_{\mathcal C}(\bar a)$ \textbf{then return} $\operatorname{Release}$}

\AppLine{0}{$c\gets(\kappa(\mathcal C),\bar a)$}

\AppLine{0}{$y\gets\operatorname{Invoke}(\mathrm{realization},x,c,I)$}

\AppLine{0}{\textbf{if} $y\neq\bot$ and $\operatorname{Preserves}(y,c,\mathcal C,I)$ \textbf{then return} $D_I(y)$}

\AppLine{0}{\textbf{if} $y=\bot$ or $\neg\operatorname{Valid}(c,\mathcal C)$ \textbf{then return} $\operatorname{Release}$}

\AppLine{0}{$f'\gets\operatorname{PublicFailure}(y,D_I(y),c)$}

\AppLine{0}{$y'\gets\operatorname{Invoke}(\mathrm{correction},x,c,y,f',I)$}

\AppLine{0}{\textbf{if} $y'\neq\bot$ and $\operatorname{Preserves}(y',c,\mathcal C,I)$ \textbf{then return} $D_I(y')$}

\AppLine{0}{\textbf{return} $\operatorname{Release}$}
\end{AppAlgorithm}

Here $\operatorname{Preserves}(y,c,\mathcal C,I)$ implements the gate in
Equation~\ref{eq:dart_full_gate}. The decoded realization must match the
committed operation and every normalized typed argument and remain valid under
the active contract. Native execution follows the shared runtime, and
$\operatorname{Release}$ returns control to the ordinary retry policy.

\subsection{Trigger Scope and Benchmark Instantiation}
\label{app:dart_current}

\DARTLead{Trigger scope.}
Across all benchmark--interface settings, DART is activated only for retryable,
policy-owned failures under a confirmed unchanged environment state. Eligible
triggers capture cases in which the current action cannot be successfully
executed or does not produce the intended progress, including interface-level
decoding or validation failures and backend execution rejections. This scope
ensures that DART intervenes only when the current commitment remains
semantically valid but requires a corrected realization. The phase-specific
prompt templates and response contracts are provided in Appendix~\ref{app:prompts}.

\DARTLead{Benchmark instantiation.}
All RQ4 settings use the same commitment-preserving controller and semantic
preservation criterion. DART operates exclusively on policy-visible task
information and public retry context, with concise task or
failure reminders included when required by the benchmark. After a commitment
is selected and validated, it is realized through the interface contract
assigned to the corresponding setting. Benchmark-specific prompt realizations
may differ in presentation, but they follow the same selection, validation, and
correction procedure described above.

\subsection{RQ4 Paired Evaluation}
\label{app:dart_final_protocol}

\DARTLead{Protocol-matched comparison.}
RQ4 evaluates Gemini-2.5-Flash in all six benchmark--interface settings. Each
DART condition is paired with a Base condition on the same task set. Within a
pair, the model, interface renderer, public task information, ordinary policy,
generation settings, semantic validator, execution pipeline, and episode-level
budgets are shared. DART adds only the error-triggered
commitment-preserving controller and the registered trigger/selection
configuration in Appendix~\ref{app:dart_current}. Endpoint score and
token cost are aggregated with the paired task-level procedure defined in
Appendix~\ref{app:metrics}.

\DARTLead{ScienceWorld protocol.}
Both ScienceWorld DART and Base runs use the same stable domain-order-v2
public-input protocol. The RQ4 comparison is therefore entirely within this
shared protocol, with DART as the controlled intervention.

\DARTLead{Cost accounting.}
All RQ4 runs use temperature zero, top-$p=1$, and the request limits specified
in Appendix~\ref{app:current_configuration}. Endpoint token cost includes every
dispatched ordinary, semantic-selection, realization, preserving-correction,
and fallback response. These requests consume the same episode-level model-call
budget used by the protocol-matched Base.
\section{Prompt Templates and Response Contracts}
\label{app:prompts}

This appendix specifies the response contracts used by the ordinary policy and
the phase-specific DART requests. Ordinary interaction follows a shared policy
instruction together with the currently rendered interface contract. When DART
activates, the same model first selects a canonical semantic action, then
realizes the validated commitment through the assigned interface, and, when
needed, performs one commitment-preserving correction. The prompts below map
directly to the controller stages defined in
Appendix~\ref{app:dart_implementation}.

\subsection{Prompt Assembly and Ordinary Policy}

Ordinary policy requests use the same message structure across MCP and CLI.
The shared system instruction in Prompt~\ref{prompt:shared} is combined with
the task instruction, current observation, rendered interface contract, visible
interaction history, and final action request, in that order. The system-level
policy and message ordering remain fixed. Interface-specific information enters
through the rendered contract, required response surface, and visible
interaction context. Outside DART trigger events, this ordinary policy governs
the trajectory unchanged.

\AppPrompt{prompt:shared}{Shared ordinary-policy instruction}{
You are the only policy acting in an interactive benchmark. Complete the task using only the task instruction, current observation, visible interaction history, and the current interface contract.

The interface contract in the user message is authoritative. Select only the single best next action for the current state; do not wait for future actions to appear, because the contract is regenerated after every accepted environment transition. Keep any internal deliberation concise and prioritize returning a complete action before the output limit. Produce exactly one response container in the requested Native, controlled CLI-style, or controlled MCP-style syntax. Do not add explanations, Markdown, aliases, extra keys, guessed parameters, or text outside that container.

An accepted action can still make no progress. If the visible response is unchanged and a controller notice identifies an immediately repeated accepted action, choose a different currently grounded action; do not repeat that action again.

Treat different complete entity names as distinct. Never substitute a shorter, longer, or merely overlapping visible name for an entity explicitly named by the task. If the exact entity is not currently visible, choose a grounded action from the current contract to navigate, inspect, or otherwise gather visible state.

If the previous proposal was rejected, use the visible validation feedback to make the single allowed correction attempt. Never assume that a rejected pre-execution proposal changed the environment.
}

\subsection{Semantic Selection Contract}

When DART activates, semantic selection replaces the interface-specific output
rule with the canonical selection contract in
Prompt~\ref{prompt:selection_system}, while preserving the same policy-visible
task and retry context. Prompt~\ref{prompt:recommitment} then requests one
canonical semantic action and is followed by the canonical output schema,
triggering response, and public failure feedback. The model must return exactly
one JSON object containing \AppLiteral{schema\_id} and
\AppLiteral{arguments}. This object is decoded and validated against the
current semantic contract, and the commitment is created only after successful
validation. Interface-specific realization begins only after this semantic
commitment has been established. The same public-schema prompting and local
validation path is used for both local and API-backed models, without guided
decoding.

\AppPrompt{prompt:selection_system}{Selection-only system output rule}{
This call is the semantic recommitment phase, not the interface-realization phase. Return exactly one JSON object with exactly schema\_id and arguments, conforming to the Canonical output schema in the user message. Do not use a Native, CLI, or MCP response container, command line, tool call, Markdown, or explanatory text for this call. This phase-specific output rule supersedes surface-format instructions and examples in the current interface contract and historical failure feedback. The current public semantic contract remains authoritative for valid operations and grounded argument values. Select one valid semantic action only; do not execute or realize it. The separate realization call will use the assigned interface syntax.
}

\AppPrompt{prompt:recommitment}{Explicit semantic recommitment}{
For this retryable realization failure only, make an explicit semantic recommitment for the current task and unchanged environment state. The failed text does not establish an original latent intent. Choose one currently valid action using the same public capabilities above. For this call only, replace the interface output format by a JSON object with exactly \AppLiteral{schema\_id} and \AppLiteral{arguments}, obeying the supplied canonical JSON schema. Do not execute or realize the action yet.
}

\subsection{Realization and Preserving Correction}

Once semantic selection has been validated, DART stores the semantic action
together with its contract identity as the commitment. The initial realization
request supplies this fixed commitment together with the assigned interface
response contract and asks the model to express exactly that action through the
required surface form. If realization fails while the commitment remains
valid, the preserving-correction request additionally supplies the failed
realization and its public failure feedback while leaving the commitment
unchanged. Prompt~\ref{prompt:realization} gives the exact instructions for
both branches. Appendix~\ref{app:rq4_current_results} provides recorded
examples of successful first-pass realization and preserving correction.

\AppPrompt{prompt:realization}{Realization and preserving-correction instructions}{
\textbf{Initial realization.} Realize the validated semantic commitment below using the assigned interface response contract above. Preserve the operation and every typed argument exactly. Do not select a different action. Return only the required interface expression.

\textbf{Preserving correction.} Correct the failed realization using the assigned interface response contract above.
Preserve the committed operation and every typed argument exactly. Correct only its interface expression.
Return only the required interface expression.

}
\endgroup

\end{document}